%% file: main.tex
\documentclass[10pt,journal,compsoc]{IEEEtran}
\usepackage[nocompress]{cite}
\usepackage[pdftex]{graphicx}
\graphicspath{{figures/}}
\DeclareGraphicsExtensions{.pdf,.jpeg,.png}
\usepackage{amsmath,bm}
\usepackage{algorithm}
\usepackage{algorithmic}
\usepackage{array}
\usepackage{url}
\usepackage[x11names]{xcolor}
\usepackage{tabulary,xfrac,enumitem}
\usepackage{booktabs}       
\usepackage{amsfonts}       
\usepackage{nicefrac}       
\usepackage{microtype}      
\usepackage{amssymb}
\usepackage{amsmath}
\usepackage{caption, subcaption}
\usepackage{tabularx}
\usepackage{multicol}
\usepackage{multirow}
\usepackage{textcomp}
\usepackage{makecell}
\usepackage{colortbl}
\usepackage{pifont}
\usepackage{xspace}
\usepackage[pagebackref=true,breaklinks=true,colorlinks,citecolor=citecolor,bookmarks=false]{hyperref}

\usepackage{xcolor}
\usepackage{color, colortbl}
\usepackage{verbatim}
\usepackage{marvosym}

\usepackage[capitalize]{cleveref}
\crefname{section}{Sec.}{Secs.}
\Crefname{section}{Section}{Sections}

\Crefname{table}{Table}{Tables}
\crefname{table}{Tab.}{Tabs.}

\newcommand{\name}{BEVHeight}
\newcommand{\nameplus}{BEVHeight++}
\newcommand{\mypara}[1]{\vspace{1mm}\noindent\textbf{#1}}

\newcommand{\extend}[1]{{\color{black}#1}}
\newcommand{\ky}[1]{{\color{black} #1}}

\definecolor{citecolor}{RGB}{34,200,34}

\makeatletter
\newcommand{\thickhline}{%
    \noalign {\ifnum 0=`}\fi \hrule height 0.5pt
    \futurelet \reserved@a \@xhline
}
\makeatother

\begin{document}
\title{BEVHeight++: Toward Robust Visual Centric 3D Object Detection}

\author{Lei~Yang,
        Tao~Tang,
        Jun~Li,
        Peng~Chen,
        Kun~Yuan,
        Li~Wang,\\
        Yi~Huang,
        Xinyu~Zhang\textsuperscript{\Letter},
        Kaicheng~Yu%
\IEEEcompsocitemizethanks{
\IEEEcompsocthanksitem L. Yang is with School of Vehicle and Mobility,  Tsinghua University, China. Email: yanglei20@mails.tsinghua.edu.cn. Part of the work done when as an intern in Cainiao Network, Alibaba Group.
\IEEEcompsocthanksitem J. Li, L. Wang, Y. Huang, and X. Zhang is with School of Vehicle and Mobility, Tsinghua University, China. Email: $\{$lijun1958, xyzhang, huangyi21@mails, wangli\_thu@mail$\}$.tsinghua.edu.cn.
\IEEEcompsocthanksitem T. Tang is with Shenzhen Campus, Sun Yat-sen University, China. Email: trent.tangtao@gmail.com.
\IEEEcompsocthanksitem P. Chen is with Cainiao Network, Alibaba Group. Email: yuanshang.cp@alibaba-inc.com.
\IEEEcompsocthanksitem K. Yuan is with Center for Machine Learning Research, Peking University, China. Email: kunyuan@pku.edu.cn.
\IEEEcompsocthanksitem K. Yu is with Westlake University, China. Email: kyu@westlake.edu.cn.
}

\thanks{Corresponding author: Xinyu Zhang.}
}

\markboth{Journal of \LaTeX\ Class Files,~Vol.~14, No.~8, August~2021}%
{Shell \MakeLowercase{\textit{et al.}}: A Sample Article Using IEEEtran.cls for IEEE Journals}

\IEEEtitleabstractindextext{%
\begin{abstract}
While most recent autonomous driving system focuses on developing perception methods on ego-vehicle sensors, people tend to overlook an alternative approach to leverage intelligent roadside cameras to extend the perception ability beyond the visual range. 
We discover that the state-of-the-art vision-centric bird's eye view detection methods have inferior performances on roadside cameras. 
This is because these methods mainly focus on recovering the depth regarding the camera center, where the depth difference between the car and the ground quickly shrinks while the distance increases. 
In this paper, we propose a simple yet effective approach, dubbed \extend{BEVHeight++}, to address this issue. In essence, we regress the height to the ground to achieve a distance-agnostic formulation to ease the optimization process of camera-only perception methods. \extend{By incorporating both height and depth encoding techniques, we achieve a more accurate and robust projection from 2D to BEV spaces.}
On popular 3D detection benchmarks of roadside cameras, our method surpasses all previous vision-centric methods by a significant margin. \extend{In terms of the ego-vehicle scenario, our BEVHeight++ possesses superior over depth-only methods. Specifically, it yields a notable improvement of +1.9\% NDS and +1.1\% mAP over BEVDepth when evaluated on the nuScenes validation set. Moreover, on the nuScenes test set, our method achieves substantial advancements, with an increase of +2.8\% NDS and +1.7\% mAP, respectively.}

\end{abstract}

\begin{IEEEkeywords}
Autonomous driving, Vision-Centric perception, Robustness, 3D object detection
\end{IEEEkeywords}
}

\maketitle
\input{latex/fig/teaser}

\IEEEdisplaynontitleabstractindextext

%
\IEEEpeerreviewmaketitle

\IEEEraisesectionheading{\section{Introduction}\label{sec:introduction}}
\input{latex/1-intro}

\section{Related Work}\label{sec:related_work}
\input{latex/2-relatedworks}

\section{Method}\label{sec:method}
\input{latex/3-method}

\section{Experiments}\label{sec:exps}
\input{latex/4-exps}

\section{Conclusion} \label{sec:conclusion}
We notice that in the domain of roadside perception, the depth difference between the foreground object and background quickly shrinks as the distance to the camera increases, this makes state-of-the-art methods that predict depth to facilitate vision-based 3D detection tasks sub-optimal. On the contrary, we discover that the per-pixel height does not change regardless of distance. To this end, we propose a simple yet effective framework, \extend{\nameplus{}}. \extend{Apart from retaining the depth pipeline. we further predict the height and project the 2D feature to bird’s-eye-view space based height prediction, Then, an accurate and height-aware BEV representation is generated by incorporating height and depth encoding techniques from both image-view and bird’s-eye-view.}
Through extensive experiments, \extend{\nameplus{}} surpasses BEVDepth by a margin of 6.27\% on DAIR-V2X-I benchmark under the traditional clean settings, and by 28.2\% on robust settings where external camera parameters change.
\extend{Comprehensive experiments on nuScenes demonstrate that our method can still maintain superiority over the depth-only methods, e.g. BEVDepth and BEVDet on ego-vehicle point of view. We hope our work can shed light on studying more effective feature representation on universal perception encompassing both roadside and ego-vehicle scenarios.}

\section*{ACKNOWLEDGMENTS} 
This work was partially supported by the National High
Technology Research and Development Program of China
under Grant No. 2018YFE0204300, the National Natural
Science Foundation of China under Grant No. 62273198,
U1964203, 52221005; and by Research Internship program of Alibaba Group.

\ifCLASSOPTIONcaptionsoff
  \newpage
\fi

\bibliographystyle{IEEEtran}
\bibliography{IEEEabrv,egbib}

\end{document}

%% file: latex/fig/teaser.tex
\begin{figure*}[!t]
\centering
\includegraphics[width=0.99\textwidth]{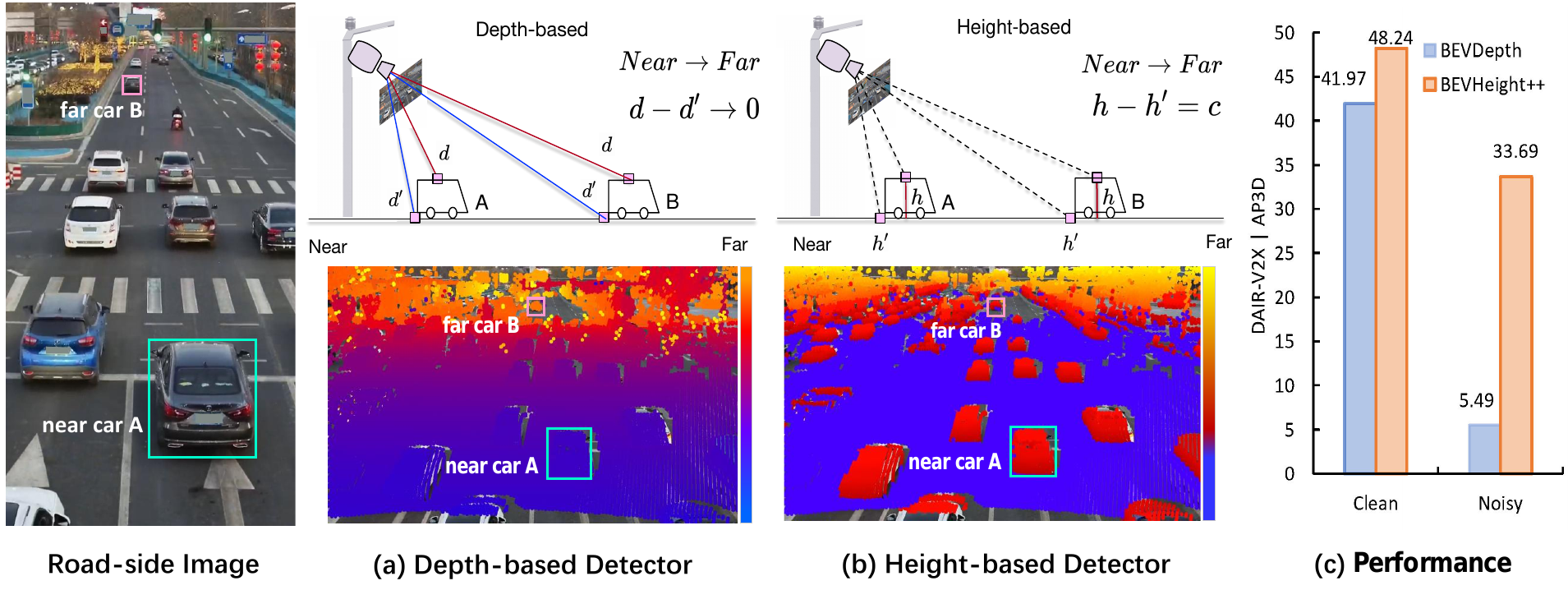}
\captionof{figure}{
\textbf{(a)} To produce 3D bounding boxes out of a monocular image, state-of-the-art methods firstly predict the per-pixel depth either explicitly or implicitly to determine the 3D location of foreground objects with the background. However, when we plot the per-pixel depth on the image, we notice that the differences between points on the car roof and surrounding ground quickly shrink when the car moves away from the camera, making it sub-optimal to optimize especially for far objects. \textbf{(b)} On the contrary, we plot the per-pixel height to the ground and observe that such difference remains agnostic regardless of the distance, and visually is superior for the network to detect objects. However, one cannot directly regress the 3D location by solely predicting the height.  \textbf{(c)} To this end, we propose a novel framework, \extend{\nameplus{}} to address this issue. Empirical results reveal that our method surpasses the best method by a margin of \extend{5.49\% on clean settings and over 28.2\% on noisy settings}.
}
\label{fig:teaser}
\end{figure*}

%% file: latex/1-intro.tex
\IEEEPARstart{T}{he} rising tide of autonomous driving vehicles draws vast research attention to many 3D perception tasks, of which 3D object detection plays a critical role.  
While most recent works tend to only rely on ego-vehicle sensors, there are certain downsides of this line of work that hinder the perception capability under given scenarios. For example, as the mounting position of cameras is relatively close to the ground, obstacles can be easily occluded by other vehicles to cause severe crash damage. To this end, people have started to develop perception systems that leverage intelligent units on the roadside, such as cameras, to address such occlusion issues and enlarge perception range so as to increase the response time in case of danger through cooperative techniques \cite{yu2023vehicle,fan2023quest,song2023spatial}. To facilitate future research, there are two large-scale benchmark datasets~\cite{yu2022dair, ye2022rope3d} of various roadside cameras and provide an evaluation of certain baseline methods. 

Recently, people have discovered that, in contrast to directly projecting the 2D images into a 3D space, leveraging a bird's eye view~(BEV) feature space can significantly improve the perception performance of vision vision-centric system. One line of the recent approach, which constitutes the state-of-the-art camera-only method, is to generate implicitly or explicitly the depth for each pixel to ease the optimization process of bounding box regression. However, as shown in \cref{fig:teaser}, we visualize the per-pixel depth of a roadside image and notice a phenomenon. Consider two points, one on the roof of a car and another on the nearest ground. If we measure the depth of these points to the camera center, namely $d$ and $d'$, the difference between these depth $d-d'$ would drastically decrease when the car moves away from the camera. We conjecture this leads to two potential downsides: i) unlike the autonomous vehicle that has a consistent camera pose, roadside ones usually have different camera poses across the datasets, which makes regressing depth hard; ii) depth prediction is very sensitive to the change of extrinsic parameter, where it happens quite often in the real world. 


On the contrary, we notice that the height to the ground is consistent regardless of the distance between the car and the camera center. \extend{Based on this motivation, we propose a novel framework to accomplish a more accurate and robust 3D object detection, dubbed \nameplus{}. Specifically, our method firstly predicts categorical height distribution for each pixel to project rich contextual feature information to the appropriate height interval in wedgy voxel space. Followed by a voxel pooling operation to get height-based BEV features, wihch is referred to as the height-based branch. 
From ~\cite{yang2023bevheight}, we observe that only relying on the height-based branch yields favorable performance on high-amounted roadside cameras. However, when it comes to ego-vehicle cameras at lower positions, it possess inferior performance compared with the depth-based method. To accomplish a more widely applicable framework, we preserve the depth-based branch in BEVDepth~\cite{li2022bevdepth} and integrate depth and height encoding techniques from both image-view and bird's-eye-view. Finally, the fused BEV features that harness the complementary advantages of depth and height representations are used as the input of detection to get output detections.}

We conduct extensive experiments on two popular roadside perception benchmarks, DAIR-V2X~\cite{yu2022dair} and Rope3D~\cite{ye2022rope3d}. On traditional settings where there is no disruption to the cameras, our \nameplus{} achieves state-of-the-art performance and surpasses all previous methods, regardless of monocular 3D detectors or recent bird's eye view methods by a margin of 5\%. In realistic scenarios, the extrinsic parameters of these roadside units can be subject to changes due to various reasons, such as maintenance and wind blows. We simulate these scenarios following \cite{yu2022benchmarking} and observe a severe performance drop of the BEVDepth, from 41.97\% to 5.49\%. Compared to these methods, we showcase the benefit of predicting the height instead of depth and achieve 26.88\% improvement over the BEVDepth~\cite{li2022bevdepth}, which further evidences the robustness of our method.
\extend{On the nuScenes dataset, our proposed method showcases a remarkable improvement of +1.9\% in NDS and +1.1\% in mAP compared to the depth-only BEVDepth when evaluated on the nuScenes validation set. Moreover, on the nuScenes test set, our method demonstrates substantial advancements, resulting in an increase of +2.8\% in NDS and +1.7\% in mAP, respectively. Specifically, when evaluated under the robust setting, considering sensor noise, motion blur, and object scale as proposed in ~\cite{dong2023benchmarking}, our framework significantly outperforms the depth-based baseline by a margin of +2.1\% to +9.4\% in mAP. This signifies the robustness and effectiveness of our approach in ego-vehicle scenarios.}

\extend{
Overall, our contributions are as follows:
\begin{itemize}
    \item We propose a novel height-based scheme that achieves great distance-agnostic representation and strong robustness to extrinsic perturbations, which is a nice replacement for complex per-pixel depth in vision-centric BEV perception.
    \item By fusing height and depth representation in both image view by concatenation and bird's-eye view through deformable cross-attention operation, we present BEVHeight++, a robust vision-centric BEV perception framework applicable to both ego-vehicle and roadside scenarios.
    \item We evaluate the proposed framework on multiple challenging benchmarks, including DAIR-V2X, Rope3D, and nuScenes. For the roadside DAIR-V2X, our \nameplus{} surpasses the best method by a margin of 6.27\% on clean settings and over 28.2\% on noisy settings. For the ego-vehicle nuScenes, our BevHeight++ outperforms the BEVDepth by  +1.9\% NDS on the validation set and achieves +2.8\% NDS on the test set.
\end{itemize}
}

%% file: latex/2-relatedworks.tex
\mypara{Roadside Perception.} Accurate environmental perception is a crucial aspect of autonomous driving and intelligent traffic systems. Concurrent perception efforts for autonomous driving are mainly limited to the ego vehicle~\cite{9409679, 9658214, 9200697, 10153690, 9983516}. However, 
\ky{as the height of vehicles are fairly low, the mounted cameras has a limited range of view and many blind spots, which limits the perception system's capability. }
Fortunately, with the rapid advancement of intelligent infrastructure, it has become possible to utilize roadside cameras for traffic environment perception. The increased height of these cameras 
\ky{leads to}
a larger field of view and long-range observation. As a result, roadside cameras have the potential to improve the accuracy and reliability of environmental perception in autonomous driving and intelligent traffic.
However, in comparison to the relatively fixed installation parameters of on-board cameras, each roadside camera has various specifications, including roll, pitch angle, and mounting height, leading to domain gaps that present additional challenges to roadside 3D detection. These gaps create ambiguities that make tasks more complex. Furthermore, while the use of roadside cameras enables the perception system to detect a wider range of obstacles, it also increases the density of obstacles, resulting in even greater complexity for the perception system.
Recently, some pioneers have presented roadside datasets~\cite{ye2022rope3d, yu2022dair,v2x-seq,cress2022a9,yu2023v2x}, hoping to facilitate the 3D perception tasks in roadside scenarios. A significant amount of research has also been conducted in the field of 3D perception for roadside applications. InfraDet3D~\cite{zimmer2023infradet3d}, for example, fuses data from multiple roadside sensor units and combines various levels of fusion (early and late fusion) along with both traditional and learning-based approaches into a unified framework. Jia et al.~\cite{jia2023competition} propose a monocular 3D object detector that utilizes a separate feature extraction module for the depth image. They then fuse the depth features and RGB features at different stages to help solve the domain gap problem of roadside images.

In order to solve the above problems and promote roadside perception to assist intelligent transportation, we take the advances and challenges of roadside cameras into account and design an efficient and robust roadside perception framework, \nameplus{}.

\mypara{Vision Centric BEV Perception.}
\extend{
Recent vision-centric works predict objects in 3D space, which is very suitable for applying multi-view feature aggregation under BEV for autonomous driving. Popular methods can be divided into the transformer-based and depth-based schema. 
Transformer-based detectors design a set of object queries~\cite{wang2022detr3d, liu2022petr, liu2022petrv2, Chen2022PolarPF, Saha2022TranslatingII} or BEV grid queries\cite{li2022bevformer,jiang2022polarformer,tseng2022crossdtr}, then perform the view transformation through cross-attention between queries and image features. DETR3D~\cite{wang2022detr3d}, for instance, begins by projecting predefined queries onto images and using an attention mechanism to model their relation with the multi-view features. BEVFormer~\cite{li2022bevformer} extends this concept further by incorporating dense queries, while PolarFormer~\cite{jiang2022polarformer} introduces polar coordinates into the model construction of the BEV space to improve performance. These methods explicitly index local image features from a 2D perspective view to 3D space, enabling better alignment between training targets and image features. Other methods employ implicit encoding of geometric information to model the view transformation, building interaction between 3D queries and image tokens. PETR~\cite{liu2022petr}, for example, eliminates the need for explicit BEV feature construction by element-wise fusing perspective feature maps with 3D positional embedding feature maps, followed by object detection using a DETR-style decoder. CrossDTR~\cite{tseng2022crossdtr} utilizes depth-guided transformers that compose depth-aware embeddings to help alleviate the false positive bounding boxes commonly existing in prior multi-view approaches.
The depth-based approaches, such as Lift-splat-shoot (LSS)~\cite{philion2020lift}, involve lifting each image into a frustum of features for each camera and then splatting these frustums into a rasterized bird's eye view (BEV) grid. Using this approach, BEVDet~\cite{huang2021bevdet} can project images explicitly into the BEV space and perform 3D object detection. Followup works introduce depth supervision from the LiDAR sensors ~\cite{li2022bevdepth} or multi-view stereo techniques ~\cite{wang2022sts, li2022bevstereo} to improve the depth estimation accuracy and achieve state-of-the-art performance. BEVDepth~\cite{li2022bevdepth} enhances the performance of BEVDet by incorporating explicit depth supervision with projected LiDAR points, resulting in state-of-the-art performance. BEVStereo~\cite{li2022bevstereo}, on the other hand, leverages temporal multi-view stereo (MVS) technology to improve the depth estimation accuracy of camera-based systems. Finally, STS~\cite{wang2022sts} proposes a novel technique that utilizes geometry correspondence to facilitate accurate depth learning.
Temporal information is crucial in recognizing occlusions and inferring the motion state of objects. To this end, BEVDet4D~\cite{huang2022bevdet4d} extends BEVDet by fusing the features of previous frames with the current frame after removing the impact of ego-motion. Similarly, SOLOFusion~\cite{park2022solofusion} aligns the BEV features from a long history of previous timesteps to the current timestep and concatenates them for long-term temporal fusion. PETRv2~\cite{liu2022petrv2}, on the other hand, extends the temporal version from PETR and directly achieves temporal alignment in 3D space based on the perspective of 3D position embeddings. Finally, StreamPETR~\cite{wang2023exploring} adopts an object-centric paradigm to propagate temporal information through object queries frame by frame, resulting in leading performance improvements at a negligible storage and computation cost.

However, when applying these methods to roadside perception, the bonus of accurate depth information fades. As the complex mounting positions and variable extrinsic parameters of the roadside cameras, and common corruption issues encountered in on-vehicle cameras, predicting depth from them is difficult. In this work, our \nameplus{} utilizes the height estimation to achieve state-of-the-art performance and best robustness.
}

%% file: latex/3-method.tex
\input{latex/fig/histogram-depth-height}
\input{latex/fig/correlation-depth-height}

We first give a brief problem definition of camera-only 3D object detection on the roadside. We then analyze the downside of predicting depth that is widely adopted in current camera-only methods and show the benefit of using height instead. Subsequently, we present our framework in detail. 

\subsection{Problem Definition}
\label{sec:problem_definition}
In this work, we would like to detect a three-dimensional bounding box of given foreground objects of interest. Formally, we are given the image $I\in R^{H\times W\times 3}$ from the roadside cameras, whose extrinsic matrix $E\in R^{3\times 4}$ and intrinsic matrix $K\in R^{3\times 3}$ can be obtained via camera calibration. 
We seek to precisely detect the 3D bounding boxes of objects on the image. We denote all bounding boxes of this image as $B=\left\{B_1,B_2,…,B_n\right\}$, and the output of detector as $\hat{B}$. 
Each 3D bounding box $B_{i}$ can be formulated as a vector with 7 degrees of freedom:
\begin{equation}
    \hat{B}_{i} = \left(x, y, z, l, w, h, \theta \right)
    \label{con:eq1}
\end{equation}
where $\left(x,y,z \right)$ is the location of each 3D bounding box . $\left(l,w,h \right)$ denotes the cuboid's length, width and height respectively. $\theta$ represents the yaw angle of each instance with respect to one specific axis. Specifically, a camera-only 3D object detector $F_{Det}$ can be defined as follows:
\begin{equation}
    \hat{B}_{ego} = F_{Det}\left(I_{cam}\right)
    \label{con:eq2}
\end{equation}
As a common assumption in autonomous driving, we assume the camera pose parameters $E$ and $K$ are known after the initial installation. In the roadside perception domain, people usually rely on multiple cameras installed at different locations to enlarge the perception range. This naturally encourages adopting those multi-view perception methods though the feature maps are not aligned geologically. Note that, although there are certain roadside units are equipped with other sensors, we focus on camera-only settings in this work for generalization purpose. 

\subsection{Comparing the depth and height}
\label{sec:delving_into_the_height}
As discussed before, state-of-the-art BEV camera-only methods first project the features into the bird's eye view space to reduce the z-axis dimension, then let the network learn implicitly \cite{liu2022petr, liu2022petrv2, li2022bevformer} or explicitly \cite{huang2021bevdet, li2022bevdepth, li2022bevstereo} about the 3D location information. Motivated by previous approaches in RGB-D recognition, one naive approach is to leverage the per-pixel depth as a location encoding.
In \cref{fig:histogram-depth-height}~(a), current methods firstly use an encoder to transform the original image into 2D feature maps. After predicting the per-pixel depth of the feature map, each pixel feature can be lifted into 3D space and zipped in the BEV feature space by voxel pooling techniques. 

However, we discover that using depth may be sub-optimal under the face-forwarding camera settings in autonomous driving scenarios. 
Specifically, we leverage the LiDAR point clouds of the DAIR-V2X-I~\cite{yu2022dair} dataset, where we first project these points to the images, to plot the histogram of per-pixel depth in \cref{fig:histogram-depth-height}~(b). We can observe a large range from 0 to 200 meters. By contrast, we plot the histogram of the per-pixel height to the ground and clearly observe the height ranges from -1 to 2m respectively, 
which is easier for the network to predict.
But in practice, the predicted height can't be employed directly to the pinhole camera model like depth. 
How to achieve the projection from 2D to 3D effectively through height has not been explored.

\subsubsection{Analysis when extrinsic parameter changes}
In \cref{fig:five}~(a), we provide an visual example of extrinsic disturbance.
To show that predicting height is superior to depth, we plot the scatter graph to show the correlation between the object's row coordinates on the image and its depth and height. Each plot represents an instance. As shown in \cref{fig:five}~(b). we observe that objects with smaller depths have a smaller $v$ value. However, suppose the extrinsic parameter changes; we plot the same metric in blue and observe that these values are drastically different from the clean setting. In that case, i.e., there is only a small overlap between the clean and noisy settings. We believe this is why the depth-based methods perform poorly when external parameters change. On the contrary, as observed in \cref{fig:five}~(c), the distribution remains similar regardless of the external parameter changes, i.e. the overlap between orange and blue dots is large. This motivates us to consider using height instead of depth. However, unlike depth that can be directly lifted to the 3D space via camera model, directly predicting height will not work to recover the 3D coordinate. Later, we present a novel height-based projection module to address this issue.

\input{latex/fig/versatility_analysis_rebuttal.tex}
\extend{\subsubsection{Limitation of height}}
\extend{
The height-based prediction error is affected by the mounting height of cameras. We visualize three cameras on different platforms observing the same object and analyze the detection error in Fig.~\ref{fig:versatility_analysis}: (a) shows when the height prediction is equal to the ground truth, detection is perfect for all cameras; (b) if not, for the same height prediction error $\lvert h_{gt} - h_{pred} \rvert$, the distance between the predicted point and ground-truth is inversely proportional to the camera ground height. Therefore, the height-based pipeline is more suitable for higher-position roadside cameras and is not friendly to the lowest-positioned ego-vehicle cameras. In order to compensate for this limitation, by fusing height and depth pipelines, we propose BEVHeight++, which is a unified vision-centric BEV perception framework applicable to both ego-vehicle and roadside scenarios.
}

\input{latex/fig/framework_bevheight_plus_plus}

\input{latex/fig/heightnet}

\extend{\subsection{BEVHeight++}}
\label{sec:BEVHeight}

\extend{\subsubsection{Network Structure}}
\extend{
As shown in Fig.\ref{fig:framework_bevheight_plus_plus}, our proposed BEVHeight++ framework consists of three sub-networks: (1) depth-based branch, (2) height-based branch, (3) feature fusion process. Firstly, the image-view encoder that is composed of a 2D backbone and an FPN module aims to extract the 2D high-dimensional multi-scale image features $F^{context} \in R^{C_c\times \frac{H}{16} \times \frac{W}{16}}$ given an image $I\in R^{3\times H \times W}$ from roadside or ego-vehicle scenarios, where $C_c$ denotes the channel number. $H$ and $W$ represent the input image's height and width, respectively. On the height-based branch, the HeightNet is responsible for predicting the bins-like distribution of height from the ground $H^{pred} \in R^{C_H \times \frac{H}{16} \times \frac{W}{16}}$, where $C_H$ stands for the number of height bins. The representations $F_{H}^{fused} $ that combines the fused features $F^{fused}$ and height distribution $H^{pred}$ is generated using Eq.~\ref{con:eq3}. The height-based $2D\rightarrow 3D$ projector pushes $F_{H}^{fused}$ into the 3D wedge-shaped features $F^{wedge} \in R^{X \times Y \times Z \times C_{f}}$ based on the predicted bins-like height distribution $H^{pred}$. See Algorithm \cref{alg:algorithm} for more details. Voxel Pooling transforms the 3D wedge-shaped features into the BEV features $F_{H}^{bev} \in R^{X \times Y \times C_{f}}$ along the height direction. On the depth-based branch, a DepthNet that estimates depth map $D^{pred} \in R^{C_D \times \frac{H}{16} \times \frac{W}{16}}$ from image features $F^{context}$, where $C_D$ stands for the number of depth bins. A View Transformer that projects $F^{fused}$ in 3D representations $F_D^{fused}$ using Eq.~\ref{con:eq4} then pools them into an integrated BEV representation $F_{D}^{bev}$. For the feature fusion process, the image-view fusion obtains fused features $F^{fused} \in R^{C_f \times \frac{H}{16} \times \frac{W}{16}}$ by concatenating height distribution $H^{pred}$ and image features $F^{context}$, which are then used by height-based and depth-based branches. The bird’s eye view fusion obtains fused BEV features $F_{fused}^{bev}$ from the height-based BEV features $F_{H}^{bev}$ and the depth-based BEV features $F_{D}^{bev} \in R^{X \times Y \times C_{f}}$. Finally, the detection head first encodes the above fused BEV features with convolution layers. And then predicts the 3D bounding box consisting of location $\left(x, y, z\right)$, dimension$\left(l, w, h\right)$, and orientation $\theta$.

\begin{equation}
    \begin{aligned}
        F_{H}^{fused} = F^{fused}\otimes H^{pred}, \\
        F_{H}^{fused} \in R^{C_f \times C_H \times \frac{H}{16} \times \frac{W}{16}}
    \end{aligned}
    \label{con:eq3}
\end{equation}

\begin{equation}
    \begin{aligned}
        F_{D}^{fused} = F^{fused}\otimes D^{pred}, \\
        F_{D}^{fused} \in R^{C_f \times C_D \times \frac{H}{16} \times \frac{W}{16}}
    \end{aligned}
    \label{con:eq4}
\end{equation}
}

\subsubsection{HeightNet}
Motivated by the DepthNet in BEVDepth~\cite{li2022bevdepth}, we leverage a Squeeze-and-Excitation layer to generate the context features  $F^{context}$ from the 2D image features $F^{2d}$. Concretely, we stack multiple residual blocks~\cite{he2016deep} to increase the representation power and then use a deformable convolution layer~\cite{XizhouZhu2018DeformableCV} to predict the per-pixel height. We denote this height module as $H^{pred}$. To facilitate the optimization process, we translate the regression task to use one-hot encoding, i.e. discretizing the height into various height bins. The output of this module is  $h\in R^{C_H\times 1\times 1}$. Moreover, previous depth discretization strategies~\cite{HuanFu2018DeepOR,YunleiTang2020Center3DCM} are generally fixed and thus not suitable for roadside height predictions. To this end, we present a dynamic-increasing discretization(DID) as follows:
\begin{equation}
    h_i =  \lfloor{N \times \sqrt[\alpha]{ \frac{h-h_{min}}{h_{max}-h_{min}}}}\rfloor,
    \label{con:eq4}
\end{equation}
where $h$ represents the continuous height value from the ground, $h_{min}$ and $h_{max}$ represent the start and end of the height range. $N$ is the number of height bins, and $h_i$ denotes the value of $i-th$ height bin. $H$ is the height of the roadside camera from the ground. $\alpha$ is the hype parameter to control the concentration of height bins. \extend{The proposed DID strategy and other existing discretization techniques are visualized in Fig. \ref{fig:discretization}}.
\input{latex/fig/space_strategy}

\subsubsection{Height-based 2D-3D projection module} 
Unlike the ``lift'' step in previous depth-based methods, one cannot recover the 3D location with only height information. To this end, we design a novel 2D to 3D projection module to push the fused features \extend{$F_{H}^{fused} \in R^{C_H \times C_f \times \frac{H}{16} \times \frac{W}{16}}$} into the wedge-shaped volume feature $F^{wedge} \in R^{X\times Y\times Z\times C_{c}}$ in the ego coordinate system.
As illustrated in \cref{fig:2d_3d_projector} and \cref{alg:algorithm}, we design a virtual coordinate system, with the origin coinciding with that of the camera coordinate system and the Y-axis perpendicular to the ground, and a special reference plane parallel to the image plane with a fixed distance 1. 

\input{latex/fig/2D_3D_projector}
\input{latex/table/alg}

For each point $p_{image} = (u, v)$ in the image plane, we first choose the associated point $p_{ref}$ in the reference plane $plane_{ref}$, whose depth is naturally 1, i.e, $d_{ref}=1$. Thus we can project $p_{ref}$ from the uvd space to the camera coordinate through the camera's intrinsic matrix:
\begin{equation}
P_{ref}^{cam}=K^{-1} d_{ref} [u,v,1]^T = K^{-1} [u,v,1]^T . 
\end{equation}
Further, it can be transformed to the virtual coordinate to get $P_{ref}^{virt.}$ with the transformation matrix $T_{cam}^{virt.}$: 
\begin{equation}
    P_{ref}^{virt.} = T_{cam}^{virt.} P_{ref}^{cam}.
\end{equation}
Now we can know the point $p_{ref}$ in our virtual coordinate is $P_{ref}^{virt.}$.
Suppose the $i-th$ value in height bins relative to the ground for point $p_{image}$ is $h_i$ and the height from the origin of the virtual coordinate system to the ground is $H$. Based on similar triangle theory, we can have the $i-th$ projected 3D point in height virtual coordinate for $p_{image}$:
\begin{equation}
P_{i}^{virt.}=\frac{H-h_i}{y_{ref}^{virt.}} P_{ref}^{virt.} .
\end{equation}
Finally, we transform the $P_{i}^{virt.}$ to the ego-car space:
\begin{equation}
P_{i}^{ego}=T_{virt.}^{ego} P_{i}^{virt.}.
\end{equation}

In summary, the contribution of our module is in two-fold: i) we design a virtual coordinate system that leverages the height from the HeightNet; ii) we adopt a reference plane to simplify the computation by setting a constant depth to 1.  We formulate the height-based 2D-3D projection as follows:
\begin{equation}
    P_{i}^{ego}=T_{virt.}^{ego} \frac{H-h_i}{y_{ref}^{virt.}} T_{cam}^{virt.} K^{-1} [u,v,1]^T.
\end{equation}

\extend{\subsubsection{Features Fusion}
\label{sec:feature_fusion}
\mypara{Image view fusion.} 
As shown in Fig. 1(b), we plot the per-pixel height on the image, and based on the height information, vehicles can be clearly distinguished from the ground. On the basis of this observation, height information is critical for 3D object detection task. However, the traditional LSS-based BEV perception methods inevitably lose height information during the voxel pooling stage, which is sub-optimal for achieving precise object detection. To this end, we propose the image view fusion before the lift operation to obtain fused features as follows:
\begin{equation}
   F^{fused} = concat(F^{context}, H^{pred})
\end{equation}

By pixel-wise coupling explicit height predictions with contextual features in the image-view, we effectively avoid the loss of height information in the voxel pooling process.

\mypara{Bird's-eye-view fusion.} 
As illustrated in Fig.~\ref{fig:versatility_analysis}, compared with the height-based pipeline, the depth-based BEV features from the depth branch tend to have smaller distance errors when approaching the camera along the direction perpendicular to the ground, which is especially evident on the ego-vehicle platform. Therefore, fusing BEV features from both the height and depth branches is essential to achieve precise object detection and provide a unified framework for both roadside and on-vehicle platforms.

When considering the BEV features $F_H^{bev}$,$F_D^{bev}$ derived from the height and depth branches, it should be noted that their error distributions differ significantly. Therefore, a simple stacking operation alone is not sufficient to achieve a high degree of alignment between features corresponding to the same spatial region. To address this problem, we design the bird’s-eye-view fusion technique as follows:

$$
F_{fused}^{bev} = BevFuse(F_H^{bev}, F_D^{bev}) =
$$

\begin{equation}
\sum_{p=0}^{HW}\left[Q'_{p} + \sum_{V \in \left\{F_H^{bev}, F_D^{bev}\right\}} DeformAttn(Q_p, p, V) \right]
\label{con:bevfusion}
\end{equation}

\noindent where $H$, $W$ are the spatial shape of BEV features, $Q_p$ denotes the BEV query located at $p=(x,y)$ of the concatenation of $F_H^{bev}$ and $F_D^{bev}$. $Q'_p$ is the sampled feature at the reference point p from the $F_H^{bev}$ in roadside scenarios or the $F_D^{bev}$ in on-vehicle scenes.
$$
DeformAttn(Q_p, p, V) =
$$
\begin{equation}
\sum_{m=1}^{M}W_m \left[\sum_{k=1}^{K} A_{mpk} \cdot W'_m V(p+\Delta p_{mpk})\right]
\label{con:deform_attn}
\end{equation}

\noindent where $M$ is the total number of attention heads, $m$ indexes the attention head, $K$ is the total number of sampled keys, and $k$ indexes the sampled keys. The attention weight $A_{mqk}$ predicted from the query feature $Q_p$ is a scalar value in the range $\left[0, 1\right]$, representing the correlation of the k-th key to the q-th query in the m-th attention head, and $\sum_{k=1}^{K}A_{mpk}=1$. $\Delta p_{mpk}$ represents the predicted offset based on the query feature $Q_p$ for the reference point p, while $V(p + \Delta p_{mpk})$ denotes the input feature at the $p + \Delta p_{mpk}$ location. $W_m \in R^{C \times C/M}$ and $W'_m\in R^{C/M \times C}$ are the learnable weights.
}

%% file: latex/fig/histogram-depth-height.tex

\begin{figure*}[h!t]
	\centering
	\includegraphics[width=1.0\textwidth]{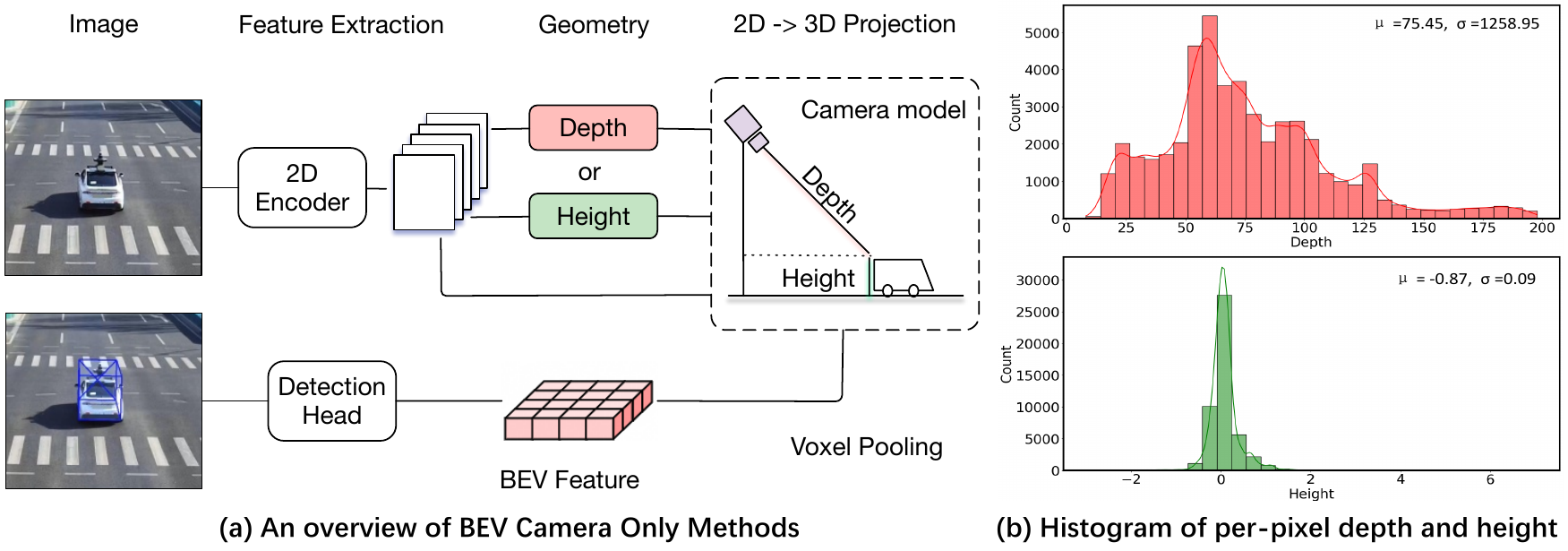}
	\caption{
	\textbf{The comparison of predicting height and depth.} 
	\textbf{(a)} We present the overview of the previous depth-based method and our proposed height-based pipeline. Note that we propose a novel 2D to 3D projection module. \textbf{(b)} We plot the histogram of per-pixel depth~(top) and ground-height~(bottom). We can clearly observe that the range of depth is over 200 meters while the height is within  5 meters, which makes height much easier to learn.
	}
\label{fig:histogram-depth-height}
\end{figure*}

%% file: latex/fig/correlation-depth-height.tex
\begin{figure*}[ht]
\centering 
\includegraphics[width=1.00\textwidth]{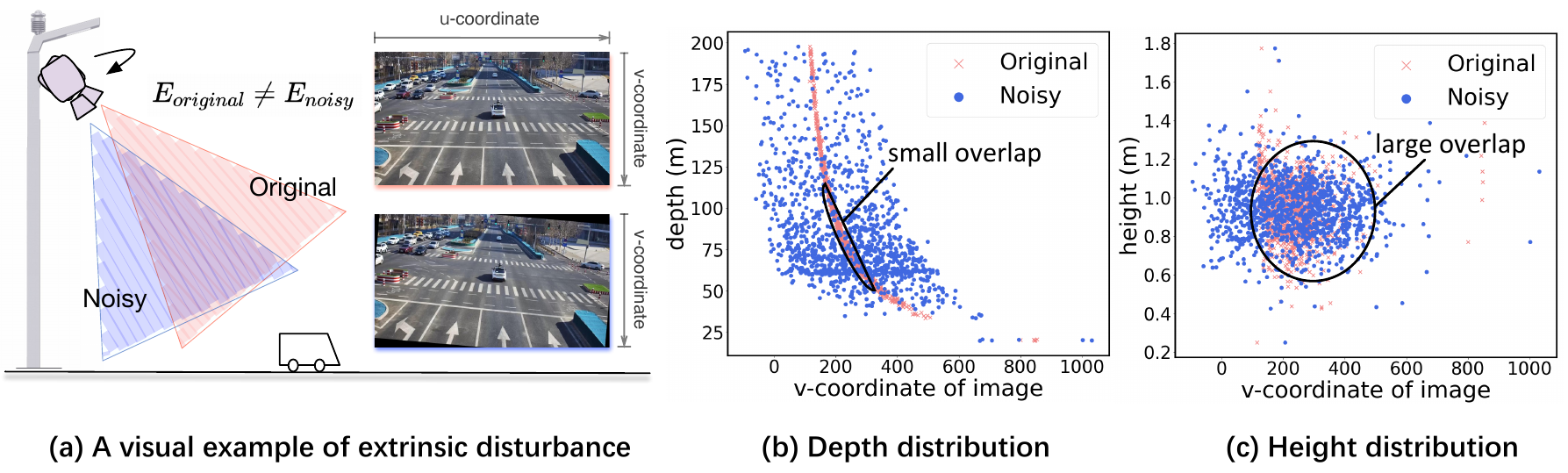}	
\caption{\textbf{The correlation between the object's row coordinates on the image with its depth and height.} The position of the object in the image, which can be defined as $(u,v)$, and $v{\raisebox{0mm}{-}}coordinate$ denotes its row coordinate of the image. (a) A visual example of the noisy setting, adding a rotation offset along roll and pitch directions in the normal distribution. (b) is the scatter diagram of the depth distribution. (c) is for the height from the ground. We can find, compared with depth, the noisy setting of height has a larger overlap with its original distribution, which demonstrates height estimation is more robust.
}

\label{fig:five}
\end{figure*}

%% file: latex/fig/versatility_analysis_rebuttal.tex
\begin{figure}[!h]
\centering
\includegraphics[width=8.5cm]{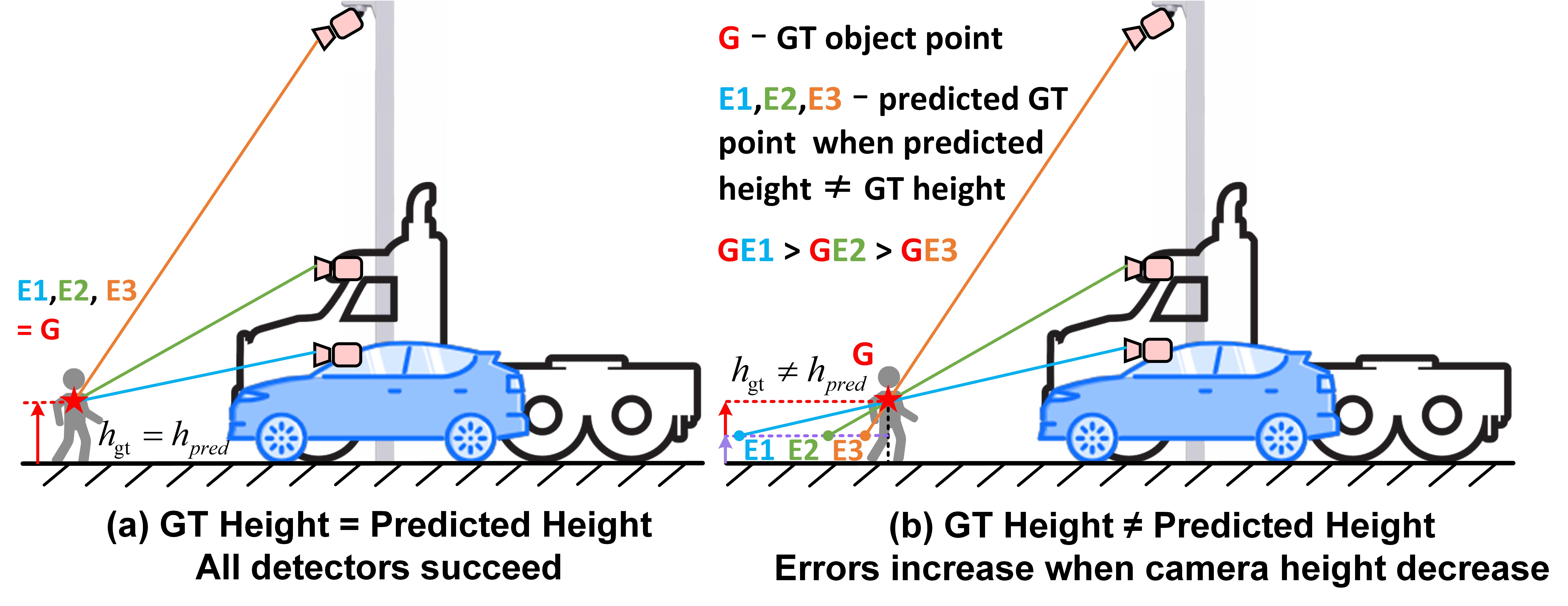}
\caption{\textbf{Distance error analysis caused by same height estimation error on different platform cameras.}}
\label{fig:versatility_analysis}
\end{figure}

%% file: latex/fig/framework_bevheight_plus_plus.tex
\begin{figure*}[t]
	\centering
	\includegraphics[width=1.0\textwidth]{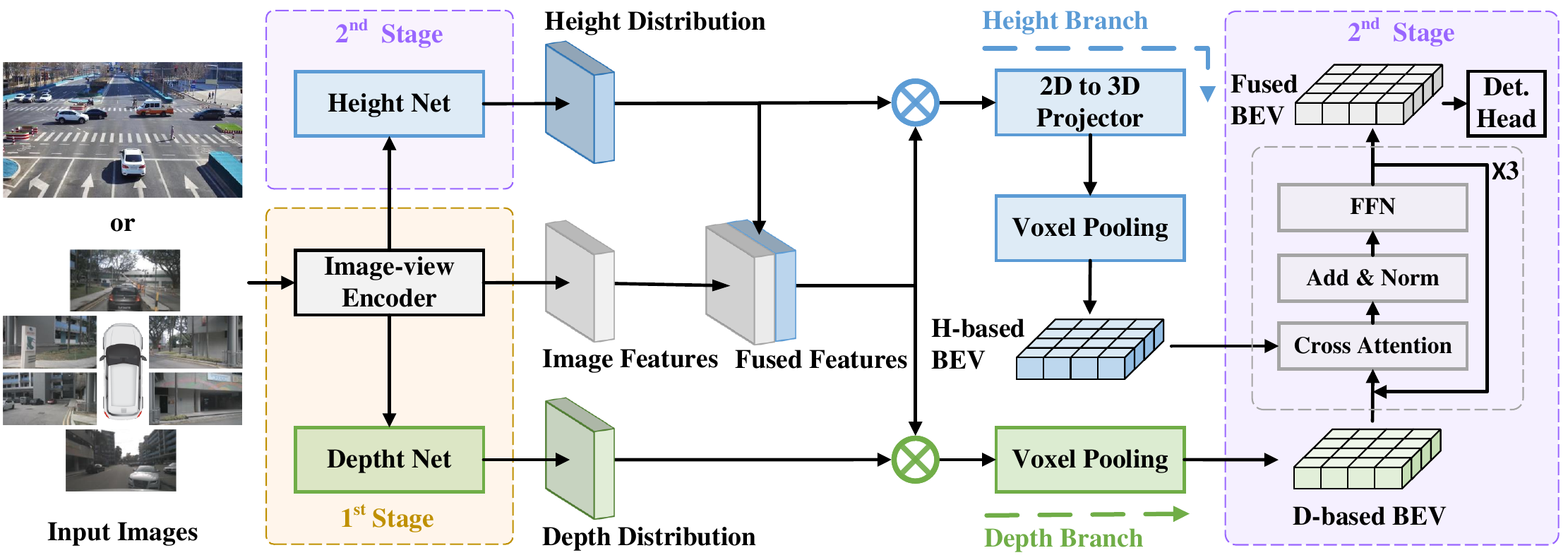}
	\caption{\extend{\textbf{The overall framework of BEVHeight++.} Our detector consists of three sub-networks for depth-based branch ({\color{cyan}{cyan}}), height-based branch ({\color{green}{green}}) and feature fusion process ({\color{gray}{gray}}). The depth-based pipeline uses the estimated per-pixel depth to lift image-view features to depth-based BEV features (D-based BEV). The height-based pipeline applies the ground-height predictions to lift features in the image view to height-based BEV features (H-based BEV). Feature fusion includes image view and bird's eye view fusion. Image-view fusion obtains fused features by concatenating height distribution and image features, which are used in the subsequent lift operation. Bird's eye view fusion obtains fused BEV features from height-based BEV features and depth-based BEV features through the deformable cross-attention, which are then used as input to the detection head.}}
\label{fig:framework_bevheight_plus_plus}
\end{figure*}

%% file: latex/fig/heightnet.tex
\begin{figure}[!h]
\centering
\includegraphics[width=0.49\textwidth]{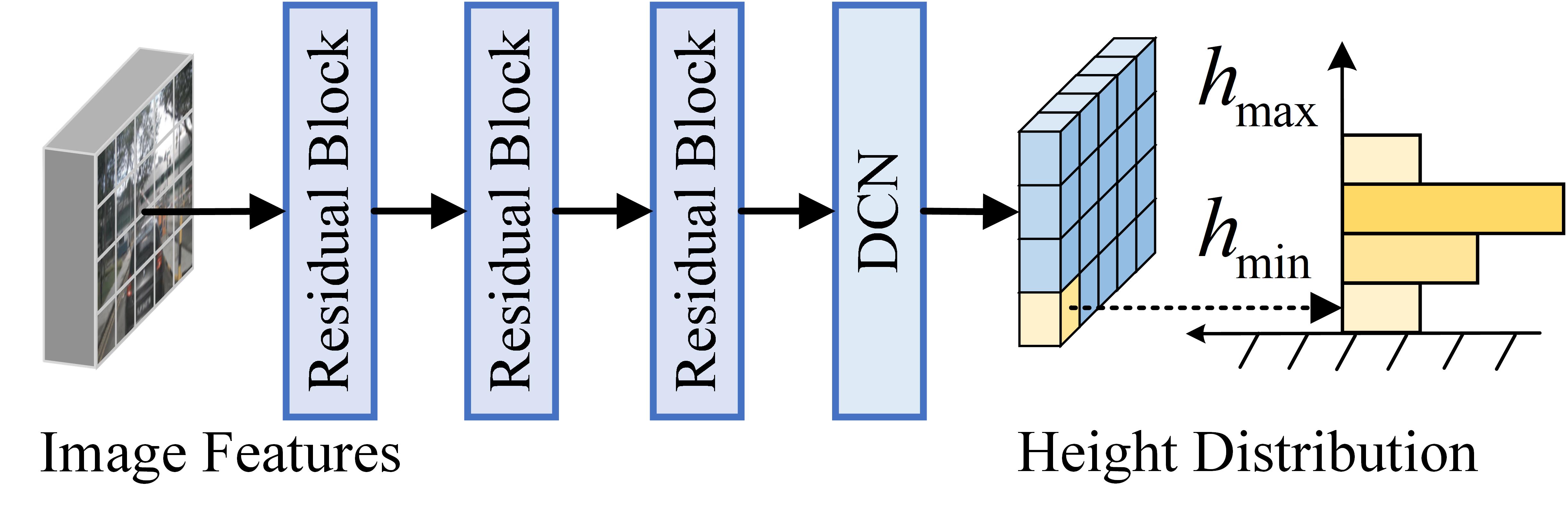}
\caption{\extend{\textbf{Framework of the height net.}
The per-pixel height distribution is predicted directly from image features. Stack multiple residual blocks and a deformable convolution layer are used for this process.
}}
\label{fig:2d_3d_projector}
\end{figure}

%% file: latex/fig/space_strategy.tex
\begin{figure}[ht]
\centering	\includegraphics[width=0.48\textwidth]{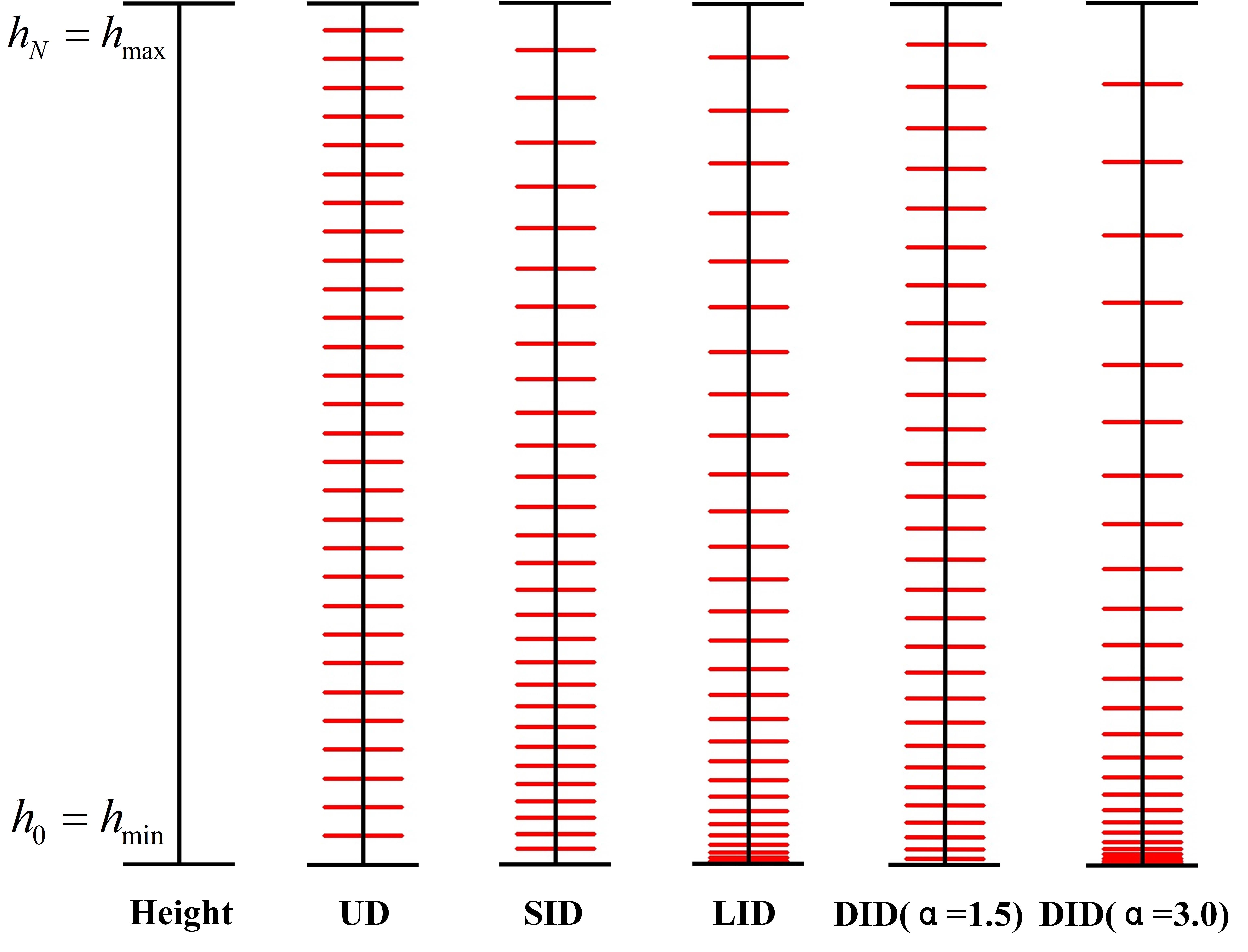}
	\caption{\extend{\textbf{Height Discretization Methods.} Height $h_i$ is discretized over a height range $[h_{min}, h_{max}]$ into $N$ discrete bins. From left to right, these are uniform discretization (UD), spacing-increasing discretization (SID), linear-increasing discretization (LID) and adjustable dynamic-increasing discretization(DID). For the dynamic-increasing discretization (DID) strategy, height bins with large $\alpha$ are more densely distributed when approaching the $h_{min}$ than the small hyper-parameter $\alpha$ conditions.}}
\label{fig:discretization}
\end{figure}

%% file: latex/fig/2D_3D_projector.tex
\begin{figure}[!h]
\centering
\includegraphics[width=0.50\textwidth]{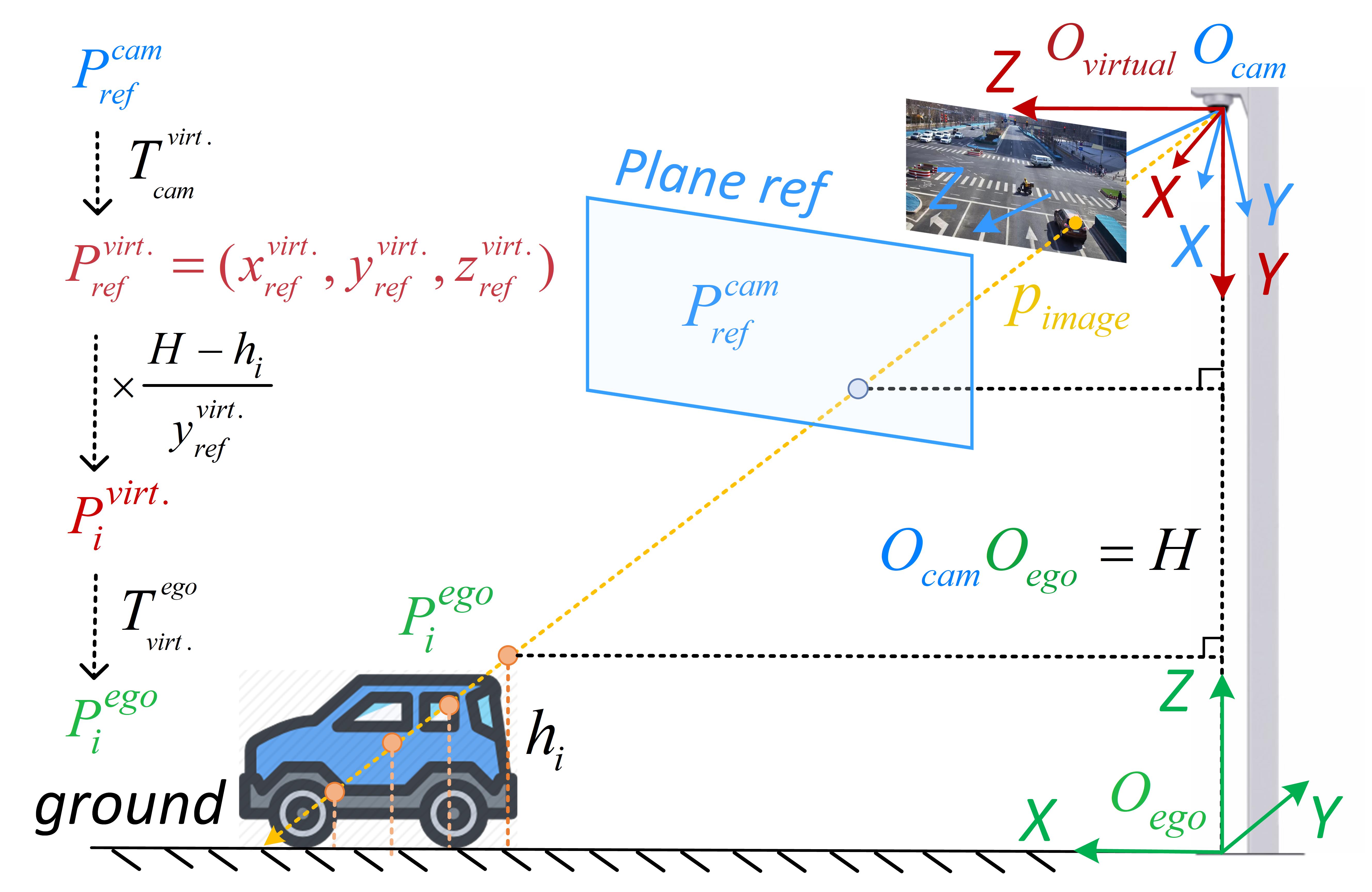}
\caption{\extend{\textbf{Height-based 2D-3D projector.} $OXYZ$ implies the coordinate system, $O_{virt.}$ has the same origin as $O_{cam}$ but with Y-axis perpendicular to the ground. A $p_{image}$ with estimated height $h_{i}$ in the image plane can be converted as a 3D point $P_{i}^{ego}$ on ego space though this 2D to 3D projection process.}}
\label{fig:2d_3d_projector}
\end{figure}

%% file: latex/table/alg.tex
\begin{algorithm}[h!t]
\caption{Height-based 2D to 3D projector}
\label{alg:algorithm}
\textbf{Parameters Definition}: \\
   $O$ and $X,Y,Z$: coordinate system, where $O_{virt.}$ has the same origin as $O_{cam}$ with Y-axis perpendicular to the ground.  \\
    $T_{A}^{B}$: transformation matrix from coordinate A to B. \\
    $K$: the camera's intrinsic matrix. \\
    $H$: the distance from the origin of the virtual coordinate system to the ground.\\
    $h_i$: the height from the ground of i-th height bin.\\
    $P_{ref}^{B}$: the pixel $(u,v)$ projected from reference plane A in coordinate B \\
    $P_i^{A}$: the pixel $(u, v)$ projection point on i-th height bin in the coordinate system A.\\
\vspace{-0.2cm}
\hrule
\vspace{0.1cm}
\textbf{Input}: \\
   \extend{$F_{H}^{fused} = \left\{f_1^{fused}, ...,f_{\frac{H}{16} \times \frac{W}{16}}^{fused}\right\}$, $f_m^{fused} \in R^{C_H \times C_f}$}\\
   $H$; $K$; $T_{cam}^{virt.}$; $T_{cam}^{ego}$\\
\textbf{Output}: \\
    $F_{wedge}$ is the 3D wedge-shaped volume features.\\
\textbf{Begin:}
\begin{algorithmic}[1] 
\STATE $F_{wedge} = \left\{\right\}$
\FOR {$f_m^{fused}$ in $F^{fused}$}
 \STATE $u, v \gets m$
 \STATE $P_{ref}^{cam}=K^{-1} [u,v,1]^T$\\
 \STATE $P_{ref}^{virt.}= \left\{x_{ref}^{virt.},y_{ref}^{virt.},z_{ref}^{virt.}\right\} = T_{cam}^{virt.} P_{ref}^{cam}$
 \FOR {$i\gets 0$ to $C_H$}
    \STATE $P_i^{virt.}=\frac{H-h_i}{y_{ref}^{virt.}}P_{ref}^{virt.}$\\
     \STATE $P_i^{ego}= T_{virt.}^{ego}P_i^{virt.}$\\
     \STATE $F_{wedge} \gets F_{wedge} \cup associate(P_i^{ego}, f_m^{fused}[i])$\\
  \ENDFOR
\ENDFOR
\STATE \textbf{return} $F_{wedge}$\\
\end{algorithmic}
\textbf{End}
\end{algorithm}

%% file: latex/4-exps.tex
We briefly introduce the experiment settings and two benchmark datasets in the roadside perception domain and one dataset in the ego-vehicle scenario. We then compare our proposed \nameplus{} with state-of-the-art methods under clean and robust settings. Experiments involving heavy ablation are carried out to confirm the efficacy and validity of \nameplus{}.


\subsection{Datasets}
\mypara{DAIR-V2X.} Yu et al.~\cite{yu2022dair} introduces a large-scale, multi-modality dataset. As the original dataset contains images from vehicles and roadside units, this benchmark consists of three tracks to simulate different scenarios. Here, we focus on the DAIR-V2X-I, which only contains the images from mounted cameras to study roadside perception. Specifically, DAIR-V2X-I contains around ten thousand images, where 50\%, 20\% and 30\% images are split into train, validation and testing respectively. However, up to now, the testing examples are not yet published, we evaluate the results on the validation set. We follow the benchmark to use the average perception of the bounding box as in KITTI~\cite{geiger2012we}.

\mypara{Rope3D\cite{ye2022rope3d}.} 
There is another recent large-scale benchmark named Rope3D. It contains over 500k images with three-dimensional bounding boxes from seventeen intersections. Here, we follow the proposed homologous setting to use 70\% of the images as training, and the remaining as testing. Note that, all images are randomly sampled. For validation metrics, we leverage the AP$_{\text{3D}{|\text{R40}}}$~\cite{simonelli2019disentangling} and the Rope$_\text{score}$, which is a consolidated metric of the 3D AP and other similarities metrics, such as average area similarity.

\extend{\mypara{nuScenes\cite{Caesar2019nuScenesAM}.} 
The purpose of this dataset is to assess the performance of autonomous driving algorithms, and it comprises data from 6 cameras, 1 LiDAR, and 5 radars. The dataset contains 1000 scenarios, divided into 700 for training, 150 for validation, and 150 for testing. The dataset includes 1.4 million annotated 3D bounding boxes for ten different object classes. The detection task is evaluated based on mean average precision (mAP) across four thresholds determined by the center distance on the ground plane. The algorithm's performance is evaluated using five true-positive metrics, including ATE for measuring translation error, ASE for scale error, AOE for orientation error, AVE for velocity error, and AAE for attribute error. Additionally, the nuScenes detection score (NDS) is defined, which combines detection accuracy (mAP) with the five true-positive metrics. The nuScenes dataset's camera frame rate is 12Hz, and image keyframes synchronized with LIDAR and RADAR are sent to annotation partners for labelling. The labelled data is used to train and evaluate the algorithms. The remaining unlabelled data comprises 10Hz raw images, which are five times more than the labelled data, and these are utilized for semi-supervised learning. Overall, the nuScenes dataset provides a comprehensive evaluation of autonomous driving algorithms, which can help advance the development of autonomous vehicles.}

\extend{\mypara{nuScens-C.~\cite{dong2023benchmarking}} Dong et al. benchmarks the robustness of 3D object detection against common corruptions in autonomous driving. This study presents 27 types of corruptions that are commonly encountered in real scenarios using LiDAR and camera sensors to comprehensively and rigorously evaluate the corruption robustness of current 3D object detectors. The corruptions are grouped into weather, sensor, motion, object, and alignment levels. By applying them to typical autonomous driving datasets KITTI\cite{geiger2012we}, nuScenes\cite{Caesar2019nuScenesAM}, and Waymo\cite{sun2020scalability}, three corruption robustness benchmarks KITTI-C, nuScenes-C, and Waymo-C are established. We mainly follow the nuScenes-C benchmark.
}

\subsection{Implementation Details}
\extend{
We conduct experiments on two setups: (1) BEVHeight, which consists of the only height-based branch and detection head. (2) BEVHeight++, which consists of the height-based branch, depth-based branch, feature fusion process, and detection head.

For architectural details, we use ResNet-101\cite{he2016deep} and V2-99\cite{Lee2019CenterMaskRA} as the 2D backbone in the results compared with state-of-the-art and ResNet-50 for other ablation studies. We set the image size to 864x1536 for the DAIR-V2X and Rope3D datasets. The input resolution for nuScenes is $512 \times 1408$ in SOTA experiments and $256 \times 704$ in ablation studies.
Following BEVDepth\cite{li2022bevdepth}, we apply various data augmentation techniques to enhance the image and BEV data. These techniques include random cropping, random scaling, random flipping, and random rotation for image data, while for BEV data we use random scaling, random flipping, and random rotation.

Our training consists of two stages: i) We first train only the depth-based branch, which is equivalent to BEVDepth. ii) We then reload and freeze the weights of the above depth-based branch and train the remaining height-based branch, feature fusion process, and detection head. We use AdamW optimizer~\cite{loshchilov2017adamw} with a learning rate $2e-4$. All models are trained on 8 V100 GPUs for 20 epochs in nuScenes dataset and 100 epochs for other roadside datasets.
}

\input{latex/table/dair_v2.tex}

\input{latex/table/rope3d}

\subsection{Quantitative Results}
\subsubsection{On \extend{roadside} benchmark}
 On DAIR-V2X-I setting, we compare our BEVHeight \extend{and BEVHeight++} with other state-of-the-art methods like ImvoxelNet~\cite{rukhovich2022imvoxelnet}, BEVFormer~\cite{li2022bevformer}, BEVDepth~\cite{li2022bevdepth} on DAIR-V2X-I val set. Some results of LiDAR-based and multimodal methods reproduced by the original DAIR-V2X~\cite{yu2022dair} benchmark are also displayed.
As can be seen from Tab.~\ref{dair_sota_2}, the proposed \extend{\nameplus{}} surpasses state-of-the-art methods by a significant margin of \extend{5.04\%, 7.46\% and 5.05\%} in vehicle, pedestrian and cyclist categories respectively.

On the Rope3D dataset, we also compare our BEVHeight \extend{and BEVHeight++} with other leading BEV methods, such as BEVFormer~\cite{li2022bevformer} and BEVDepth~\cite{li2022bevdepth}. Some results of the monocular 3D object detectors are revised by adapting the ground plane. As shown in Tab.~\ref{tab_performance_overall},  
we can see that our method outperforms all BEV and monocular methods listed in the table. In addition, under the same configuration, our \nameplus{} outperforms the BEVDepth by \extend{$6.49\%$ / $5.09\%$, $6.21\%$ / $5.28\%$} on AP$_{\text{3D}{|\text{R40}}}$ and Rope$_\text{score}$  for car and big vehicle respectively.

\input{latex/table/nuscenes_val_set}
\input{latex/table/nuscenes_test_set}
\input{latex/table/nuscenes_corruption_map}

\input{latex/table/nuscenes_corruption_nds}

\extend{\subsubsection{On ego-vehicle benchmark}}
\extend{On the nuScenes\cite{Caesar2019nuScenesAM} benchmark, we evaluated our BEVHeight++ and compared it with several state-of-the-art BEV detectors including BEVFormer~\cite{li2022bevformer}, PolarFormer~\cite{jiang2022polarformer}, PETRv2~\cite{liu2022petrv2}, BEVDet4D~\cite{huang2022bevdet4d}, BEVDepth~\cite{li2022bevdepth}, BEVStereo~\cite{li2022bevstereo}, Fast-BEV~\cite{Li2023FastBEVAF}, and SOLOFusion~\cite{Wang2023ExploringOT}. Tab.~\ref{tab:nus_val_performance} and Tab.~\ref{tab:nus_test_performance} show the 3D object detection results on the nuScenes validation set and test set, respectively. For the validation set in Tab.~\ref{tab:nus_val_performance}, our proposed method outperforms existing advanced methods in terms of mAP and NDS. For example, when ResNet-50 is used as the backbone, BEVHeight++ achieves 37.3\% mAP and 49.8\% NDS, which is significantly outperforming the BEVDepth-R50 [31] baseline by +2.2\% mAP and +2.3\% NDS. In addition, with the larger backbone ResNet-101 and larger image resolution, BEVHeight++ achieves the best performance of 55.4\% NDS, surpassing BEVDet4D-Swin-T~\cite{huang2022bevdet4d} 51.5\% NDS, BEVDepth-R101~\cite{li2022bevformer} 53.5\% NDS, Fast-BEV-R101~\cite{Li2023FastBEVAF} 53.5\% NDS and SOLOFusion-R101~\cite{Wang2023ExploringOT} 54.4\% NDS. This improvement in overall performance can be attributed to our proposed height-based pipeline and feature fusion module, which effectively improve the 3D spatial representation of the existing depth-based pipeline.
For the test leaderboard in Tab.~\ref{tab:nus_test_performance}, our BEVHeight++ with V2-99 backbone outperforms all entries on the nuScenes camera-only 3D objection leaderboard with 62.8\% NDS and 52.0\% mAP, surpassing the second-place method SOLOFusion by +0.9\% on NDS metric. This significant improvement highlights the significant benefits of our proposed height-based scheme.
}

\extend{On the nuScenes-C~\cite{dong2023benchmarking} benchmark, We compare the corruption robustness of our method with other detectors on the nuScenes-C dataset. We mainly focus on common corruption issues encountered in cameras, including object-level distortions such as scale, shear, motion-level corruptions like moving objects, blur, and sensor-level distortions like gaussian noise and impulse noise. The results for the mAP metric are presented in Tab.~\ref{tab:nuscenes_c_map}, while the results for the NDS metric are presented in Tab.~\ref{tab:nuscenes_c_nds}. As depicted in Tables \ref{tab:nuscenes_c_map} and \ref{tab:nuscenes_c_nds}, let's take scale corruption at the object level as an illustrative example. It is apparent that the performance of existing methods experiences a significant degradation. Specifically, there is a substantial decline in the mean Average Precision (mAP) of FCOS3D~\cite{Wang2021FCOS3DFC}, PGD~\cite{wang2022probabilistic}, and BEVDepth~\cite{li2022bevdepth}, decreasing from 23.8\% to 6.8\%, from 23.2\% to 6.6\%, and from 41.2\% to 19.1\%, respectively. In comparison with these aforementioned methods, our \nameplus{} consistently maintains a 28.5\% mAP under the same conditions. This result is particularly remarkable as it outperforms the BEVDepth baseline by +9.4\%, thereby demonstrating its substantial robustness against common corruptions.
}

\subsubsection{On robust settings}
In the realistic world, camera parameters frequently change for various reasons. Here we evaluate the performance of our framework in such noisy settings. We follow \cite{yu2022benchmarking} to simulate the scenarios in which external parameters are changed. Specifically, we introduce a random rotational offset in normal distribution $N(0, 1.67)$ along the roll and pitch directions as the mounting points usually remain unchanged.  
During the evaluation, we add the rotational offset along roll and pitch directions to the original extrinsic matrix. The image is then applied with rotation and translation operations to ensure the calibration relationship between the new external reference and the new image. Examples are given in Sec.~\ref{sec:visualization_results}.
As shown in Tab.~\ref{dair_robust}, the performance of the existing methods degrades significantly when the camera's extrinsic matrix is changed. Take $\text{Vehicle}_{(IoU=0.5)}$ for example, the accuracy of BEVFormer~\cite{li2022bevformer} drops from 50.73\% to 16.35\%. The decline of BEVDepth~\cite{li2022bevdepth} is from 60.75\% to 9.48\%, which is pronounced. Compared with the above methods, Our \extend{\nameplus{}} maintains \extend{53.46\%} from the original \extend{65.52\%}, which surprises the BEVDepth by \extend{43.98\%} on the vehicle category.
\input{latex/table/disturb}

\subsection{Qualitative Results}
\label{sec:visualization_results}
\subsubsection{On DAIR-V2X-I Dataset.}
As shown in Fig.~\ref{fig:visualization}, we present the results of BEVDepth~\cite{li2022bevdepth} and our \nameplus{} in the image view and BEV space, respectively. The above two models are not applied with data augmentations in the training phase. From the samples in (a), we can see that the predictions of BEVHeight fit more closely to the ground truth than that of BEVDepth. As for the results in (b), under the disturbance of roll angle, there is a remarkable offset to the far side relative to the ground truth in BEVDepth detections. In contrast, the results of our method are still keeping the correct position with ground truth.  Moreover, referring to the predictions in (c), BEVDepth can hardly identify far objects, but our method can still detect the instances in the middle and long-distance ranges and maintain a high IoU with the ground truth.
\subsubsection{On nuScenes Dataset.}
We present the visualized detection results for both image-view and bird’s-eye-view perspectives in Fig.~\ref{fig:visualization_ego}. In contrast to BEVDepth~\cite{li2022bevdepth}, \nameplus{} yields more precise predictions by leveraging a height-based pipeline. For instance, the green dashed rectangles demonstrate that \nameplus{} exhibits a substantial enhancement in detecting distant, occluded vehicles. Meanwhile, the orange dashed rectangles signify the benefits of our approach in detecting smaller objects.

\input{latex/fig/visualization}
\input{latex/fig/visualization_ego}

\section{Ablation Study}
\subsection{Ablation Studies on BEVHeight}
\subsubsection{Dynamic Discretization}
Experiments in Tab.~\ref{dair_discretization} show the detection accuracy improvement 0.3\% - 1.5\% when our dynamic discretization is applied instead of uniform discretization(UD).
The performance when hype-parameter $\alpha$ is set to 2.0 suppresses that of 1.5 in most cases, which signifies that hype-parameter $\alpha$ is necessary to achieve the most appropriate discretization.
\input{latex/table/discretization}

\input{latex/table/bevdet_rebuttal.tex}

\input{latex/table/latency_rebuttal.tex}

\input{latex/table/dair2_v2}

\input{latex/fig/distance_correlation}

\subsubsection{Effectiveness on multi-depth-based Detectors} 
We extend our modules on BEVDepth\cite{li2022bevdepth} and BEVDet~\cite{huang2021bevdet} on
 DAIR-V2X-I\cite{yu2022dair} and present the results here. As shown in Tab. \ref{tab:bevdet_rebuttal}, replacing the depth-based projection in BEVDepth\cite{li2022bevdepth}, our method achieves a performance increase of 2.19\%, 5.87\%, 4.61\% on vehicle, pedestrian and cyclist. Similarly, our approach surpasses the origin BEVDet by 8.56\%, 5.35\%, 8.60\% respectively.

 \subsubsection{Latency}
As shown in Tab.~\ref{latency_rebuttal}, we benchmark the runtime of BEVHeight and BEVDepth. With an image size of 864x1536, BEVDepth runs at 14.7 FPS with a latency of 68ms, while ours runs at 16.1 FPS with 62ms, which is around 5\% faster. It is due to the depth range (1$\sim$104m) being much larger than height (-1$\sim$1m), thus ours has 90 height bins that are less than 206 depth ones,
leading to a slimmer regression head and fewer pseudo points for voxel pooling. It evidences the superiority of predicting height instead of depth and advocates the efficiency of our method.

\subsubsection{More Results on DAIR-V2X-I Dataset}
Tab.~\ref{dair} shows the experimental results of deploying our proposed approach on the DAIR-V2X-I\cite{yu2022dair} val set. Under the same configurations (e.g., backbone and BEV resolution), our model outperforms the BEVDepth\cite{li2022bevdepth} baselines by a large margin, which demonstrates the admirable performance of our approach.

\subsubsection{Analysis on Distance Error}
To provide a qualitative analysis of depth and height estimations, we convert depth and height to the distance between the predicted object's center and the camera’s coordinate origin, as is shown in Fig.~\ref{fig:distance_correlation}.  Compared with the distance error triggered by depth estimation in BEVDepth\cite{li2022bevdepth}, the height estimation in our BEVHeight introduces less error, which illustrates the superiority of height estimation over the depth estimation in the roadside scenario.

\subsubsection{Analysis on Point Cloud Supervision}
\input{latex/table/pc_sup}
To verify the effectiveness of point cloud supervision in roadside scenes, we conduct ablation experiments on both BEVDepth~\cite{li2022bevdepth} and our method. As shown in Tab.~\ref{pc_sup}, BEVDepth with point cloud supervision is slightly lower than that without supervision. As for our BEVHeight, although there is a slight improvement under the IoU=0.5 condition, the overall gain is not apparent. This can be explained by the fact that the background in roadside scenarios is stable. These background point clouds fail to provide adequate supervised information and increase the difficulty of model fitting.

\extend{
\subsection{Ablation Studies on BEVHeight++}

\subsubsection{Effect of feature fusion}
We conducted experiments to investigate the impact of our proposed feature fusion process, which combines image view and bird's eye view fusion. The results, as shown in Tab.~\ref{tab:ablation_acculate_feature_fusion}, indicate that each component alone leads to a significant performance improvement over the BEVDepth~\cite{li2022bevdepth} baseline. The best performance is achieved when both components are used together. These results support our analysis in Sec.~\ref{sec:feature_fusion}, which suggests that the estimated height to the ground can effectively improve the 3D spatial representation of the existing depth-based pipeline.

\input{latex/table/ablation_accumulate_feature_fusion}

\subsubsection{The feature combinations in image-view fusion}
In this part, we study the effect of all possible feature combinations in the image-view fusion. As shown in Tab.~\ref{tab:ablation_image_view_fusion}, compared with the one only using the image features, fusing the height distribution and image features through concatenation boosts the performance in mAP and NDS by about +0.6\% and +0.9\%. However, Adding depth distribution fails to achieve additional performance improvements, which verifies the motivation that the height distribution is more critical for 3D object detection in Sec.~\ref{sec:feature_fusion}. 
\input{latex/table/ablation_image_view_fusion}

\subsubsection{The sampled features in bird's-eye-view fusion} 
In Tab.~\ref{tab:ablation_sampled_features}, we ablate the effect of the sampled features at the reference point from the height-based BEV features or depth-based BEV features. For the nuScenes\cite{Caesar2019nuScenesAM} in on-vehicle scenes, the best performance is achieved when the sampled features come from $Q_{D}^{bev}$. However, for the roadside DAIR-V2X-I~\cite{yu2022dair}, using sampled features from $Q_{H}^{bev}$ achieves higher accuracy.
\input{latex/table/ablation_sampled_features}

\subsubsection{Effect of multi-stage training}
We conduct ablation experiments to validate the effect of the multi-stage training scheme in Tab.~\ref{tab:multi_training_stage}. (a) means training all model weights at once. (b) represents first training the depth-based branch in BEVHeight++, then freezing the weights of the depth-based branch and training the remaining height-based branch, the feature fusion process, and the detection head. (c) follows the same two-stage training strategy as (b), but swaps the position of depth-based branch and height-based branch. The results in (b) outperforms the baseline (a) by +0.9\% in mAP and +0.6\% in NDS, which implies the effectiveness of the multi-stage training strategy. The results in (b) and (c) are equivalent, indicating that the selection of the first trained branch has no effect on the performance.
\input{latex/table/ablation_multi_stage_training}
\input{latex/table/ablation_versatility}

\subsubsection{Effectiveness on multi-depth-based Detectors}
We extend our height-based branch and feature fusion process on BEVDepth\cite{li2022bevdepth} and BEVDet~\cite{huang2021bevdet} on nuScenes\cite{Caesar2019nuScenesAM} validation set and present the results here. As shown in Tab.~\ref{tab:ablation_versatility}, replacing the depth-based projection in BEVDepth\cite{li2022bevdepth}, our method achieves a performance increase of +2.3\% in mAP and +2.1\% in NDS. Similarly, our approach surpasses the origin BEVDet by +2.2\% mAP and +2.3\% NDS respectively.
}

%% file: latex/table/dair_v2.tex
\begin{table}[h!t]
 \scriptsize\centering\addtolength{\tabcolsep}{-3.8pt}
\caption{\textbf{Comparing with the state-of-the-art on the DAIR-V2X-I val set.} Here, we report the results of three types of objects, vehicle~(veh.), pedestrian~(ped.) and cyclist~(cyc.). Each object is categorized into three settings according to the difficulty defined in ~\cite{yu2022dair}. First, recent BEVDepth surpasses the previous best by a large margin, showing that using bird's-eye-view indeed helps in roadside scenarios. Our method outperforms the BEVDepth by over 3\% in average precision and constitutes state-of-the-art. It is surprising to see that our method outperforms those relying on LiDAR modality.
}

 \begin{tabularx}{1.0\linewidth}{l|c|ccc|ccc|ccc}
  \toprule
 \multirow{3}{*}{Method} &  
 \multirow{3}{*}{M}  
 & \multicolumn{3}{c|}{$\text{Veh.}_{(IoU=0.5)}$} & \multicolumn{3}{c|}{$\text{Ped.}_{(IoU=0.25)}$} & \multicolumn{3}{c}{$\text{Cyc.}_{(IoU=0.25)}$} \\
    \cmidrule(r){3-11}
     &  & Easy & Mid & Hard & Easy & Mid & Hard & Easy & Mid & Hard  \\
\midrule
PointPillars~\cite{lang2019pointpillars} & L &63.07 & 54.00 & 54.01 & 38.53 & 37.20 & 37.28 & 38.46 & 22.60 & 22.49 \\
SECOND~\cite{yan2018second} & L &71.47 & 53.99 & 54.00 & 55.16 & 52.49 & 52.52 & 54.68 & 31.05 & 31.19 \\
MVXNet~\cite{Sindagi2019MVX} & LC &71.04 & 53.71 & 53.76 & 55.83 & 54.45 & 54.40 & 54.05 & 30.79 & 31.06 \\
\midrule
ImvoxelNet~\cite{rukhovich2022imvoxelnet} &C & 44.78 & 37.58 & 37.55 & 6.81 & 6.746 & 6.73 & 21.06 & 13.57 & 13.17 \\
BEVFormer~\cite{li2022bevformer} & C 	&	61.37&	50.73&	50.73&	16.89&	15.82&	15.95	&22.16&	22.13&	22.06\\
BEVDepth~\cite{li2022bevdepth}&	C 	&	
75.50&	63.58&	63.67&	34.95&	33.42&	33.27& 55.67&	55.47&	55.34\\
\midrule
 \rowcolor{cyan!30} BEVHeight & C &	
 77.78&	65.77&	65.85&	41.22&	39.29&	39.46	&60.23&	60.08&	60.54\\
 \rowcolor{cyan!30}  \extend{\nameplus} & \extend{C} &	
 \extend{\textbf{79.31}}&	\extend{\textbf{68.62}}&	\extend{\textbf{68.68}}&	\extend{\textbf{42.87}}&	\extend{\textbf{40.88}}&	\extend{\textbf{41.06}}	&\extend{\textbf{60.76}}&	\extend{\textbf{60.52}}&	\extend{\textbf{61.01}}\\
\bottomrule
\multicolumn{8}{l}{\scriptsize{M, L, C denotes modality, LiDAR, camera respectively.}}
  \end{tabularx}
  \label{dair_sota_2}
\end{table}

%% file: latex/table/rope3d.tex
\begin{table}[h!t]
\footnotesize
  \centering\addtolength{\tabcolsep}{-3.6pt}
\caption{\textbf{Results on the Rope3D val set.} Here, we follow~\cite{ye2022rope3d} to report the results on vehicles. Our method on average surpasses the state-of-the-art method over a margin of \extend{5\%} in both average precision and $Rope_{score}$ metric.
}
\vspace{-0.2cm}
 \begin{tabularx}{1.\linewidth}{ l |cc|cc|cc|cc }
\toprule
\multirow{4}{*}{Method}   & \multicolumn{4}{c|}{IoU = 0.5} & \multicolumn{4}{c}{IoU = 0.7} \\ 
\cmidrule(r){2-9}
  & \multicolumn{2}{c|}{Car} & \multicolumn{2}{c|}{Big Vehicle} & \multicolumn{2}{c|}{Car} & \multicolumn{2}{c}{Big Vehicle} \\ 
\cmidrule(r){2-9}
&AP & Rope &
AP & Rope &
AP & Rope &
AP & Rope \\

\midrule

M3D-RPN~\cite{brazil2019m3d} 
&54.19 & 62.65	&33.05 &  44.94 &16.75 & 32.90 &6.86  &  24.19 \\

Kinematic3D~\cite{brazil2020kinematic}  &50.57  & 58.86&	37.60&  48.08 &17.74  & 32.9 &   6.10&   22.88\\

MonoDLE~\cite{ma2021delving} 
 & 51.70 & 60.36 & 40.34 & 50.07  & 13.58 & 29.46 &9.63 &25.80\\

MonoFlex~\cite{zhang2021objects} & 60.33 & 66.86&	37.33 &47.96   & 33.78 & 46.12 &  10.08 &26.16\\

BEVFormer~\cite{li2022bevformer}	&50.62&	58.78&	34.58&	45.16&	24.64&	38.71	&10.05&	25.56\\

BEVDepth~\cite{li2022bevdepth}	&69.63&	74.70&	45.02&	54.64&	42.56&	53.05	&21.47	&35.82\\

\midrule
 \rowcolor{cyan!30} BEVHeight & 74.60& 78.72& 48.93& 57.70& 45.73& 55.62& 23.07& 37.04 \\	
  \rowcolor{cyan!30} \extend{\nameplus} & \extend{\textbf{76.12}}& \extend{\textbf{80.91}}& \extend{\textbf{50.11}}& \extend{\textbf{59.92}}& \extend{\textbf{47.03}}& \extend{\textbf{57.77}}& \extend{\textbf{24.43}}& \extend{\textbf{39.57}} \\

\bottomrule
\multicolumn{9}{l}{\footnotesize{AP and Rope denote AP$_{\text{3D}{|\text{R40}}}$ and Rope$_\text{score}$ respectively.}}
\end{tabularx}
\label{tab_performance_overall}
\end{table}

%% file: latex/table/nuscenes_val_set.tex
\begin{table*}[h!t]
\centering
\caption{\extend{\textbf{Comparison on the nuScenes \emph{val} set.} ``L'' denotes LiDAR, ``C'' denotes camera and ``D'' denotes Depth/LiDAR supervision. 
$\dagger$ denotes initialization from an FCOS3D\cite{Wang2021FCOS3DFC}  backbone.
}}
\begin{tabular}{l|c|c|c|cc|ccccc}
\toprule\noalign{\smallskip}
\renewcommand\arraystretch{1.20}
\textbf{Methods} & \textbf{Backbone} & \textbf{Image Size} &  \textbf{Modality} & \textbf{NDS$\uparrow$} & \textbf{mAP$\uparrow$} & \textbf{mATE$\downarrow$} & \textbf{mASE$\downarrow$} & \textbf{mAOE$\downarrow$} & \textbf{mAVE$\downarrow$} & \textbf{mAAE$\downarrow$} \\
\noalign{\smallskip}
\hline
\noalign{\smallskip}
CenterPoint-Voxel~\cite{Yin2020Centerbased3O}  & &      -   & L & 0.648    & 0.564 &   -   &    -   &     -    &     -  &    -     \\
CenterPoint-Pillar~\cite{Yin2020Centerbased3O} &  &     -        & L    & 0.602     & 0.503 &   -    &    -   &     -    &  -     &     -   \\ 
\noalign{\smallskip}
\hline
\noalign{\smallskip}
FCOS3D~\cite{Wang2021FCOS3DFC}   & R101-DCN & 900$\times$1600& C    & 0.415      & 0.343 & 0.725 & 0.263 & 0.422   & 1.292 & \textbf{0.153} \\
DETR3D$\dagger$~\cite{wang2022detr3d} & R101-DCN &900$\times$1600 & C   & 0.422      & 0.347 & 0.765 & 0.267 & 0.392   & 0.876 & 0.211  \\
DETR4D$\dagger$~\cite{Luo2022DETR4DDM} & R101-DCN &640$\times$1600 & C   & 0.509      & 0.422 & 0.688 & 0.269 & 0.388   & 0.496 & 0.184  \\
PETR$\dagger$~\cite{liu2022petr} & R101-DCN &900$\times$1600 & C   & 0.442      & 0.370 & 0.711 & 0.267 & 0.383   & 0.865 & 0.201  \\
PETRv2$\dagger$~\cite{liu2022petrv2} & R101-DCN &640$\times$1600 & C   & 0.524      & 0.421 & 0.681 & 0.267 & 0.357   & 0.377 & 0.186  \\
PolarFormer$\dagger$~\cite{jiang2022polarformer}   & R101-DCN& 900$\times$1600  & C  & 0.528    & 0.432 & 0.648 & 0.270 & 0.348   & 0.409 & 0.201 \\
BEVFormer$\dagger$ ~\cite{li2022bevformer}    & R101-DCN  &        900$\times$1600       & C    & 0.517     & 0.416 & 0.673 & 0.274 & 0.372   & 0.394 & 0.198  \\
BEVDet ~\cite{huang2021bevdet}  & Swin-T    & 512$\times$1408      & C   & 0.417      & 0.349 & 0.637 & 0.269 & 0.490   & 0.914 & 0.268 \\
BEVDet4D ~\cite{huang2022bevdet4d}  & Swin-T  & 640$\times$1600      & C    & 0.515       & 0.396 & 0.619 & 0.260 & 0.361   & 0.399 & 0.189\\
Fast-BEV~\cite{Li2023FastBEVAF} & R101 & 900$\times$1600& C & 0.535 &  0.413 & 0.584 & 0.279 & 0.311 & 0.329 & 0.206  \\
SOLOFusion~\cite{Wang2023ExploringOT}  & R101 & 512$\times$1408& C & 0.544 &  \textbf{0.472} & \textbf{0.518} & 0.275 & 0.604 & 0.310 & 0.210  \\
\noalign{\smallskip}
\hline
\noalign{\smallskip}
BEVDepth~\cite{li2022bevdepth}   &  R50 & 256$\times$704       & C\&D    & 0.475     & 0.351 & 0.639 & 0.267 & 0.479   & 0.428 & 0.198  \\
\rowcolor{cyan!30} \nameplus & R50 & 256$\times$704& C\&D & 0.498&  0.373&  0.614&  0.269& 0.419& 0.375& 0.203  \\
BEVDepth~\cite{li2022bevdepth}   & R101  & 512$\times$1408      & C\&D    & 0.535     & 0.412 & 0.565 & 0.266 & 0.358   & 0.331 & 0.190  \\
\rowcolor{cyan!30} \nameplus & R101 & 512$\times$1408& C\&D & \textbf{0.554}&  0.423& 0.541& \textbf{0.262}& \textbf{0.307}& \textbf{0.277}& 0.187\\

\bottomrule
\end{tabular}
\label{tab:nus_val_performance}
\arrayrulecolor{black}
\end{table*}

%% file: latex/table/nuscenes_test_set.tex
\begin{table*}[h!t]
\centering
\addtolength{\tabcolsep}{1.5pt}
\caption{\extend{\textbf{Comparison on the nuScenes \emph{test} set.} ``L'' denotes LiDAR, ``C'' denotes camera and ``D'' denotes Depth/LiDAR supervision. $\dagger$ indicates that V2-99\cite{Lee2019CenterMaskRA} was pretrained on the depth estimation task with extra data.
}}
\resizebox{1.0\textwidth}{!}{
\begin{tabular}{l|c|c|c|cc|ccccc}
\toprule
\textbf{Method}  & \textbf{Backbone} & \textbf{Image Size} & \textbf{Modality} & \textbf{NDS}$\uparrow$ & \textbf{mAP}$\uparrow$  & \textbf{mATE}$\downarrow$ & \textbf{mASE}$\downarrow$  & \textbf{mAOE}$\downarrow$ & \textbf{mAVE}$\downarrow$ & \textbf{mAAE}$\downarrow$ \\
\midrule
CenterPoint \cite{Yin2020Centerbased3O}   &                     & & L      & 0.648   & 0.564 & - & - & -   & - & -  \\ 
\midrule
FCOS3D~\cite{Wang2021FCOS3DFC}        &      R101-DCN                 &$900 \times 1600$   & C        & 0.428 & 0.358 & 0.690 & 0.249 & 0.452   & 1.434 & 0.124  \\
DETR3D~\cite{wang2022detr3d}   &V2-99$\dagger$ &$900 \times 1600$ & C       & 0.479 & 0.412 & 0.641 & 0.255 & 0.394   & 0.845 & 0.133  \\
PETR~\cite{liu2022petr} & V2-99$\dagger$ &$640 \times 1600$ &  C        & 0.481 & 0.434 & 0.641 & 0.248 & 0.437   & 0.894 & 0.143   \\
UVTR~\cite{li2022unifying} & V2-99$\dagger$ &$640 \times 1600$ &  C        & 0.551 & 0.472 & 0.577 & 0.253 & 0.391   & 0.508 & 0.123 \\
BEVDet4D~\cite{huang2022bevdet4d}                                 & Swin-B & $900 \times 1600$ & C  & 0.569  & 0.451 & 0.511 & 0.241 & 0.386   & 0.301 & 0.121  \\
BEVFormer~\cite{li2022bevformer}                                & V2-99$\dagger$ &$900 \times 1600$ & C   & 0.569      & 0.481 & 0.582 & 0.256 & 0.375   & 0.378 & 0.126 \\ 
PolarFormer~\cite{jiang2022polarformer}                                & V2-99$\dagger$ &$900 \times 1600$ & C   & 0.572    & 0.493 & 0.556 & 0.256 & 0.364   & 0.440 & 0.127 \\ 
PETRv2~\cite{liu2022petrv2}                           &  V2-99$\dagger$   &$640 \times 1600$ & C    & 0.582    & 0.490 & 0.561 & 0.243 & \textbf{0.361}   & 0.343 & \textbf{0.120}   \\ 
BEVStereo~\cite{li2022bevstereo}                        &  V2-99$\dagger$   &$640 \times 1600$    & C    & 0.610    & 0.525 & 0.431 & 0.246 & 0.358 & 0.357 & 0.138  \\
SOLOFusion~\cite{Wang2023ExploringOT} & ConvNeXt-B &$900 \times 1600$ & C  & 0.619 & \textbf{0.540} & 0.453 & 0.257 & 0.376 & \textbf{0.276} & 0.148 \\
\midrule
BEVDepth ~\cite{li2022bevdepth}             &    V2-99$\dagger$     &$640 \times 1600$         & C\&D  & 0.600      & 0.503 & 0.445 & 0.245 & 0.378 & 0.320 & 0.126  \\
\rowcolor{cyan!30} \nameplus{}      &    V2-99$\dagger$ &$640 \times 1600$  & C\&D        & \textbf{0.628}& 0.520& \textbf{0.295}& \textbf{0.239}& 0.365& 0.289& 0.136 \\
\bottomrule
\end{tabular}}
\label{tab:nus_test_performance}
\arrayrulecolor{black}

\end{table*}

%% file: latex/table/nuscenes_corruption_map.tex
\begin{table}[t]
\centering
\addtolength{\tabcolsep}{1.3pt}
\caption{\extend{\textbf{Comparison on the nuScenes-C robust benchmark under the mAP metric.}}}
\begin{tabular}{l|c|cc|cc}
\toprule\noalign{\smallskip}
\renewcommand\arraystretch{2.0}
\multirow{2}{*}{Method} & \multirow{2}{*}{Clean}  & \multicolumn{2}{c|}{Object} & \multicolumn{2}{c}{Motion} \\
\cmidrule(r){3-6}
~& ~& Scale & Shear & Blur & Moving \\
\noalign{\smallskip}
\hline
\noalign{\smallskip}
FCOS3D~\cite{Wang2021FCOS3DFC}  & 0.238 & 0.068  & 0.172  & 0.102 & 0.104 \\
PGD~\cite{wang2022probabilistic} & 0.232 & 0.066  & 0.167   & 0.096 & 0.105 \\
DETR3D~\cite{wang2022detr3d} & 0.347 & 0.12  & 0.175 & 0.111 & 0.166  \\
BEVFormer~\cite{li2022bevformer} &  0.417 & 0.176  & 0.247  & 0.197 & 0.202  \\
BEVDepth~\cite{li2022bevdepth} & 0.412& 0.191  & 0.272   & 0.213 & 0.219 \\
BEVHeight++ & \textbf{0.423} & \textbf{0.285}  & \textbf{0.312}   & \textbf{0.245} & \textbf{0.240} \\
\bottomrule
\end{tabular}
\label{tab:nuscenes_c_map}
\arrayrulecolor{black}
\end{table}

%% file: latex/table/nuscenes_corruption_nds.tex
\begin{table}[t]
\centering
\addtolength{\tabcolsep}{1.5pt}
\caption{\extend{\textbf{Comparison on the nuScenes-C robust benchmark under the NDS metric.
}}}
\begin{tabular}{l|c|cc|cc}
\toprule\noalign{\smallskip}
\renewcommand\arraystretch{2.0}
\multirow{2}{*}{Method} & \multirow{2}{*}{Clean}  & \multicolumn{2}{c|}{Object} & \multicolumn{2}{c}{Motion} \\
\cmidrule(r){3-6}
~& ~ & Scale & Shear & Blur & Moving \\
\noalign{\smallskip}
\hline
\noalign{\smallskip}
FCOS3D~\cite{Wang2021FCOS3DFC}  & 0.347 &0.216  & 0.293   & 0.230 & 0.235 \\
PGD~\cite{wang2022probabilistic} & 0.350 &   0.214&  0.406  & 0.235 & 0.243 \\
DETR3D~\cite{wang2022detr3d} & 0.422 &  0.254&  0.287& 0.234 & 0.281 \\
BEVFormer~\cite{li2022bevformer} & 0.517 &   0.328&  0.387 &  0.291 & 0.345 \\
BEVDepth~\cite{li2022bevdepth} & 0.535 & 0.358  & 0.403  & 0.311 & 0.364  \\
BEVHeight++ & \textbf{0.554} & \textbf{0.401} & \textbf{0.448}   & \textbf{0.355} & \textbf{0.401} \\
\bottomrule
\end{tabular}
\label{tab:nuscenes_c_nds}
\arrayrulecolor{black}
\end{table}

%% file: latex/table/disturb.tex
\begin{table}[h!t]
 \scriptsize\centering\addtolength{\tabcolsep}{-3.00pt}
 \renewcommand\arraystretch{1.10}
   \caption{
    \textbf{Results on robustness settings. } Here, we simulate the robustness scenarios where the external parameters of the camera change. Consider the  Specifically, we consider two degrees of freedom mutation, roll and pitch of the camera center. In both dimensions, we randomly sample angles from a normal distribution of $\mathcal{N}(0, 1.67)$. Surprisingly, given such minor changes, traditional depth-based methods decrease to under 15\% even for those vehicles under easy settings. On the contrary, our methods achieve around 577\% improvement compared to those baselines, evidencing the robustness of \nameplus{}.
   }
 \begin{tabularx}{1.0\linewidth}{l|cc|ccc|ccc|ccc}
  \toprule
 \multirow{3}{*}{\rotatebox{90}{Model}} & \multicolumn{2}{c|}{Disturbed}
    \ & \multicolumn{3}{c|}{$\text{Veh.}_{(IoU=0.5)}$} & \multicolumn{3}{c|}{$\text{Ped.}_{(IoU=0.25)}$} & \multicolumn{3}{c}{$\text{Cyc.}_{(IoU=0.25)}$}  \\
   \cmidrule(r){2-12}
   & roll &	pitch & Easy & Mid & Hard & Easy & Mid & Hard & Easy & Mid & Hard  \\
  
    \midrule
 \multirow{4}{*}{\rotatebox{90}{BEVFormer}} &&&	61.37&	50.73	&50.73&	16.89&	15.82&	15.95	&22.16&	22.13&	22.0\\
	&	\checkmark& 	&	50.65&	42.9&	42.95&	10.16&	9.41&	9.47&	13.62&	13.71&	13.08\\
	&		&\checkmark&	46.40&	38.26&	38.37&	9.12	&8.44&	8.55&	8.99 &	8.43&	8.42 \\
 	&	\checkmark&	\checkmark&	19.24&	16.35&	16.47&	3.93&	3.43&	3.52&	4.93&	4.98&	4.98\\
\midrule
 \multirow{4}{*}{\rotatebox{90}{BEVDepth}}	& & &			71.56& 	60.75&	60.85&	21.55&	20.51&	20.75&	40.83	&40.66&	40.26\\
&	\checkmark	&&	34.82&	28.32&	28.35&	4.49&	4.36&	4.39&	10.48&	9.51&	9.73\\
	&&	\checkmark&	14.04&	11.41&	11.49&	3.01&	2.67&	2.75&	6.43&	6.23&	6.83\\
	&	\checkmark &\checkmark &	11.84&	9.48&	9.54&	2.16&	1.84&	1.89&	4.31&	4.14&	4.26\\
\midrule
 \multirow{4}{*}{\rotatebox{90}{BEVHeight}}	&	&&	75.58&	63.49&	63.59&	26.93&	25.47&	25.78&	47.97	& 47.45	& 48.12	\\
	&	\checkmark &&	66.06&	54.99&	55.14&	18.66&	17.63&	17.78&	34.45&	26.93&	27.68\\
	&&	\checkmark&	68.49&	56.98&	57.11&	17.94&	16.87&	17.09&	34.48&	27.82&	28.67\\
	&	\checkmark &	\checkmark&	62.64&	51.77&	51.9&	14.38&	14.01&	14.09&	31.28&	25.24&	26.02\\
\midrule
 \multirow{4}{*}{\rotatebox{90}{\extend{BEVHeight++}}}	&	& &	\extend{76.98}&	\extend{65.52}&	\extend{65.64}&	\extend{27.19}&	\extend{26.88}&	\extend{27.08}&	\extend{48.14}& 	\extend{48.11}& \extend{48.63}	\\
	&	\extend{\checkmark} &&	\extend{69.45}&	\extend{57.61}&	\extend{57.76}&	\extend{19.91}&	\extend{18.94}&	\extend{19.08}&	\extend{35.77}&	\extend{27.79}&	\extend{27.97}\\
	&&	\extend{\checkmark}&	\extend{71.80}&	\extend{59.66}&	\extend{59.87}&	\extend{18.47}&	\extend{17.57}&	\extend{17.78}&	\extend{36.14}&	\extend{28.15}&	\extend{28.67}\\
	&	\extend{\checkmark} &	\extend{\checkmark}&	\extend{64.96}&	\extend{53.46}&	\extend{53.62}&	\extend{15.96}&	\extend{15.13}& \extend{15.25}&	\extend{32.25}&	\extend{25.69}&	\extend{26.93}\\

    \bottomrule
  \end{tabularx}
  \label{dair_robust}
\end{table}

%% file: latex/fig/visualization.tex
\begin{figure*}[h!t]
	\centering
	\includegraphics[width=\textwidth]{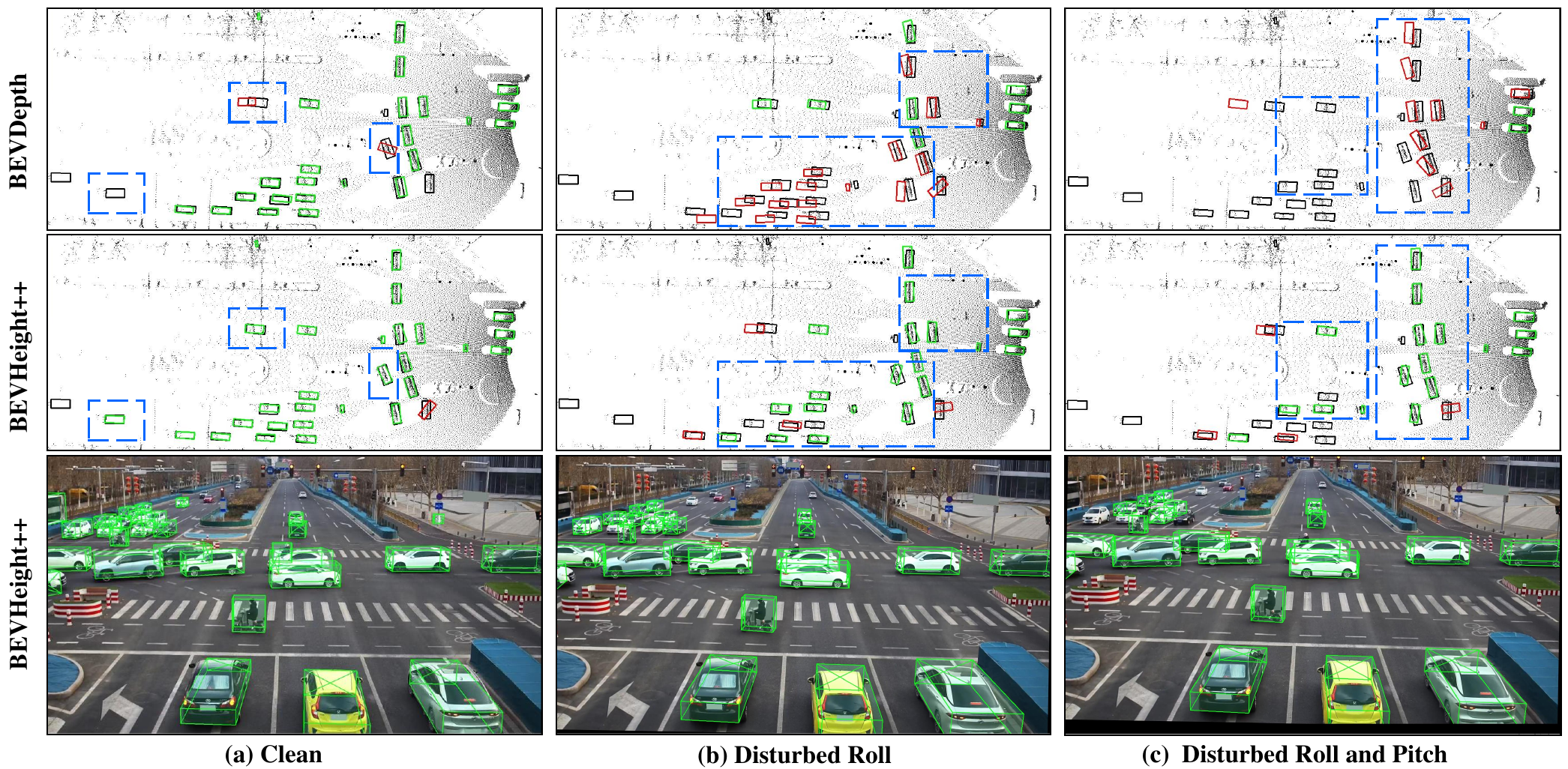}
	\caption{\textbf{Visualization results of BEVDepth and our proposed \nameplus{} under the extrinsic disturbance on roadside scenarios.} We use boxes in \textbf{{\color{red}red}} to represent false positives,  \textbf{{\color{green}green}} boxes for truth positives, and \textbf{{\color{black}black}} for the ground truth. The truth positives are defined as the predictions with IoU\textgreater 0.5 for vehicle and IoU\textgreater 0.25 for pedestrian and cyclist. (a) Clean means the original image without any processing; (b) Disturbed Roll denotes camera rotate 1 degree along roll direction; (c) Disturbed Roll and Pitch represents camera rotate 1 degree along roll and pitch directions simultaneously. 
    We observe that, our methods outperform the baseline in all three settings. Note that, the BEVDepth only identify two objects under roll-pitch disturbance while ours identify seventeen. }

\label{fig:visualization}
\end{figure*}

%% file: latex/fig/visualization_ego.tex
\begin{figure*}[h!t]
	\centering
	\includegraphics[width=\textwidth]{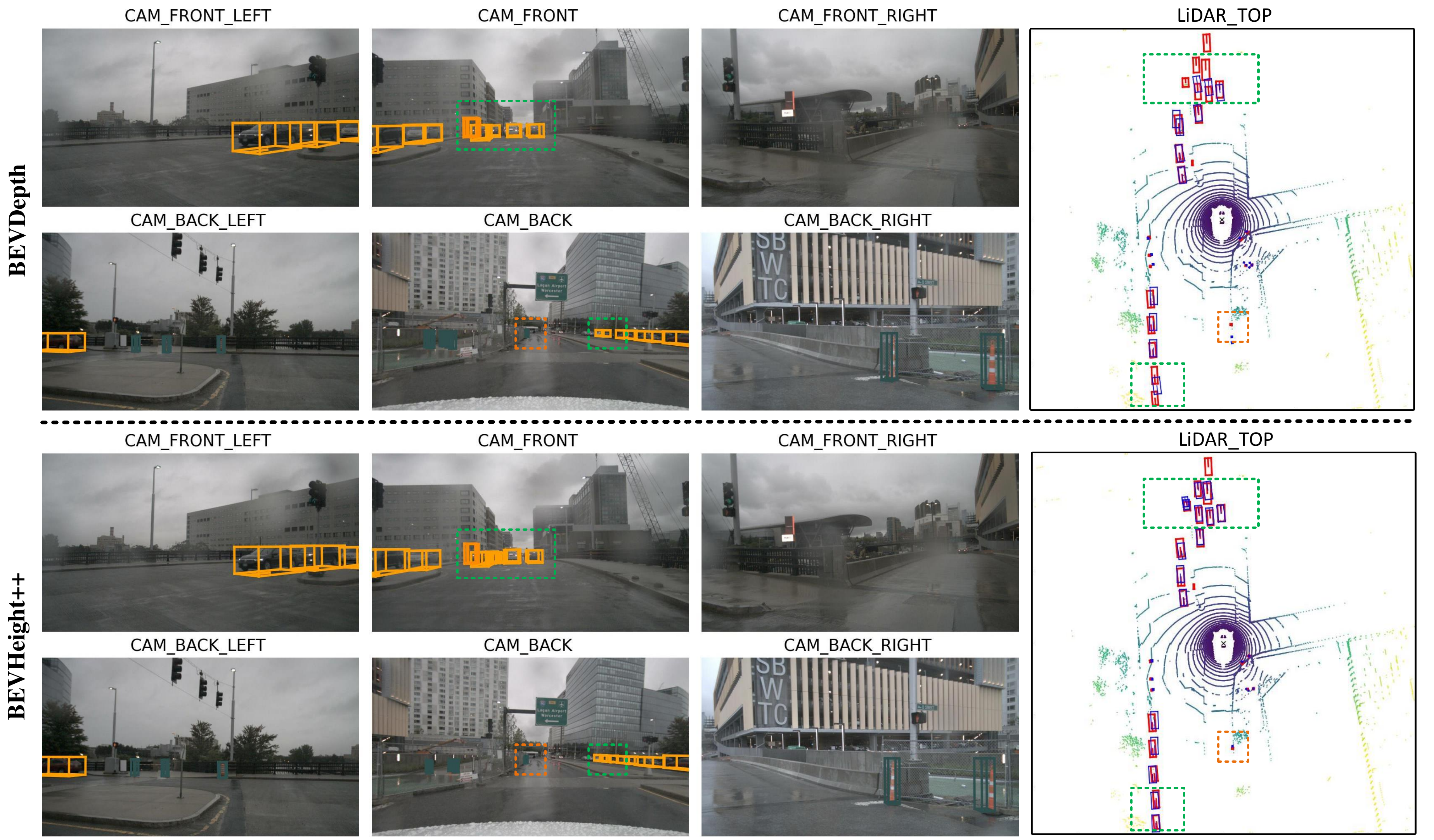}
\caption{\textbf{Visualization results of BEVDepth and \nameplus{} on ego-vehicle scenarios.} The red boxes and blue boxes on BEV represent the ground truth and the predicted boxes.}

\label{fig:visualization_ego}
\end{figure*}

%% file: latex/table/discretization.tex
\begin{table}[h!t]
 \scriptsize\centering\addtolength{\tabcolsep}{-2.7pt}
 \caption{\textbf{Ablating our dynamic discretization on DAIR-V2X-I dataset.} Compared to the uniform discretization(UD), our method achieves on average 1\% improvement in average precision. 
 }
 \begin{tabularx}{1.0\linewidth}{cc|ccc|ccc|ccc}
  \toprule
   \multicolumn{2}{c|}{Spacing}
   & \multicolumn{3}{c|}{$\text{Veh.}_{(IoU=0.5)}$} & \multicolumn{3}{c|}{$\text{Ped.}_{(IoU=0.25)}$} & \multicolumn{3}{c}{$\text{Cyc.}_{(IoU=0.25)}$} 
\\
   \cmidrule(r){1-11}
    DID ($\alpha$) &	UD & Easy & Mid & Hard & Easy & Mid & Hard & Easy & Mid & Hard  
   \\
    \midrule
   	~ & \checkmark &	75.63&	63.75&	63.85&	25.82&	25.47&	25.35&	47.52&	47.47&	47.19\\
   	\checkmark (1.5) &	~&	76.24&	64.54&	64.13&	26.47&	25.79&	25.72&	48.55&	48.21&	47.96\\
     \checkmark (2.0)& ~&	\textbf{76.61}&	\textbf{64.71}&	\textbf{64.76}&	\textbf{27.34}&	\textbf{26.09}&	\textbf{25.33}&	\textbf{49.68}&	\textbf{48.84} & \textbf{48.58}\\
    \bottomrule
  \end{tabularx}
  \label{dair_discretization}
\end{table}

%% file: latex/table/bevdet_rebuttal.tex
\begin{table}[h!t]
\scriptsize\centering
\addtolength{\tabcolsep}{-4.4pt}
\caption{\textbf{Ablation studies on different depth-based methods.} Here, we conduct the evaluation on DAIR-V2X-I val set, and report the results of three types of objects, vehicle~(veh.), pedestrian~(ped.) and cyclist~(cyc.). \extend{`VT' denotes view transformation, `D', `H' represents depth-based and height-based ones, `DH' implies the one fusing depth and height simultaneously.}}
 \begin{tabularx}{1.0\linewidth}{l|c|ccc|ccc|ccc}
 \toprule
 \multirow{3}{*}{Method} &
 \multirow{3}{*}{VT} &
\multicolumn{3}{c|}{$\text{Veh.}_{(IoU=0.5)}$} & \multicolumn{3}{c|}{$\text{Ped.}_{(IoU=0.25)}$} & \multicolumn{3}{c}{$\text{Cyc.}_{(IoU=0.25)}$} \\
 \cmidrule(r){3-11}
   &  & Easy & Mod. & Hard & Easy & Mod. & Hard & Easy & Mod. & Hard  \\
 \midrule

 \multirow{2}{*}{BEVDepth\cite{li2022bevdepth}} & D &	75.50&	63.58&	63.67&	34.95&	33.42&	33.27& 55.67&	55.47&	55.34 \\
  & {\cellcolor{cyan!30} H}&	{\cellcolor{cyan!30} 77.78}&	{\cellcolor{cyan!30} 65.77}&	{\cellcolor{cyan!30} 65.85}& {\cellcolor{cyan!30} 41.22}&	{\cellcolor{cyan!30} 39.29}&	{\cellcolor{cyan!30} 39.46}&  {\cellcolor{cyan!30} 60.23}&  {\cellcolor{cyan!30} 60.08}& {\cellcolor{cyan!30} 60.54} \\
\midrule
 \multirow{2}{*}{BEVDet~\cite{huang2021bevdet}} & D &  59.59& 	51.92&	51.81&  12.61& 12.43& 12.37& 34.91& 34.32& 34.21 	\\
& {\cellcolor{cyan!30} H}&	{\cellcolor{cyan!30} 69.42}&	{\cellcolor{cyan!30} 60.48}&	{\cellcolor{cyan!30} 59.68}& {\cellcolor{cyan!30} 18.11}&	{\cellcolor{cyan!30} 17.81}&	{\cellcolor{cyan!30} 17.74}&  {\cellcolor{cyan!30} 44.69}&  {\cellcolor{cyan!30} 42.92}& {\cellcolor{cyan!30} 42.34} \\
\bottomrule
\end{tabularx}
\label{tab:bevdet_rebuttal}
\end{table}

%% file: latex/table/latency_rebuttal.tex
\begin{table}[h!t]
\scriptsize\centering\addtolength{\tabcolsep}{-2.0pt}
\renewcommand\arraystretch{1.0}
\caption{{\bf Latency of BEVHeight and BEVDepth.} }
\begin{tabular}{l|c|c|c|c|c}
\toprule   
Methods& Backbone &Range & Number of bins & Latency (ms) & FPS \\ 
\midrule
BEVDepth~\cite{li2022bevdepth} & R50 & 1 - 104m& 206& 82& 12.2\\
\rowcolor{cyan!30} BEVHeight& R50 & -1 - 1m&  90& 77& 13.0\\
\midrule
    BEVDepth~\cite{li2022bevdepth} &  R101& 1 - 104m& 206& 68& 14.7\\
\rowcolor{cyan!30}	BEVHeight  & R101& -1 - 1m&  90& 62& 16.1\\
\bottomrule 
\multicolumn{6}{l}{\scriptsize Measured on a V100 GPU. Image shape 864×1536.}
\end{tabular}
\label{latency_rebuttal}
\end{table}

%% file: latex/table/dair2_v2.tex
\begin{table*}[h!t]
 \small 
 \centering
 \addtolength{\tabcolsep}{1.4pt}
 \caption{\textbf{Comparison on the DAIR-V2X-I Detection Benchmark.} Here, we report the results of three types of objects: Vehicle, Pedestrian and Cyclist. Each object is categorized into three settings according to the difficulty defined in ~\cite{yu2022dair}. Our BEVHeight manages to surpass the BEVDepth baseeline over a margin of 2\% - 6\% under the same configurations.}
 \begin{tabularx}{1.0\textwidth}{l|cc|ccc|ccc|ccc}
  \toprule
 \multirow{4}{*}{Method} & \multicolumn{2}{c|}{\multirow{2.8}{*}{Scale of Detector}} & \multicolumn{9}{c}{AP3D}\\
    \cmidrule(r){4-12}
    & &  & \multicolumn{3}{c|}{$\text{Vehicle}_{(IoU=0.5)}$} & \multicolumn{3}{c|}{$\text{Pedestrian}_{(IoU=0.25)}$} & \multicolumn{3}{c}{$\text{Cyclist}_{(IoU=0.25)}$} \\
   \cmidrule(r){2-12}
   & Backbone & BEV & Easy & Middle & Hard & Easy & Middle & Hard & Easy & Middle & Hard\\  
    \midrule
     BEVDepth\cite{li2022bevdepth} &	R50&	128x128	&	73.05&	61.32&	61.19&	22.10	&21.57	&21.11	&42.85&	42.26&	42.09\\
    BEVDepth\cite{li2022bevdepth} &	R101&	128x128	&	74.81&	62.44&	62.31&	24.49	&23.33&	23.17&	44.93&	44.02&	43.84\\
    BEVDepth\cite{li2022bevdepth} &	R101&	256x256	&	75.50&	63.58&	63.67&	34.95&	33.42&	33.27& 55.67&	55.47&	55.34\\
    \midrule
    BEVHeight&	R50&	128x128	&	76.61	&64.71	&64.76&	27.34&	26.09&	26.33	&49.68&	48.84&	48.58\\
    BEVHeight&	R101&	128x128	&	76.93&	64.97&	65.03&	28.53&	27.15&	27.48& 51.39&	50.83	&50.44\\
    BEVHeight& R101&	256x256	&	\textbf{77.78}&	\textbf{65.77}&	\textbf{65.85}&	\textbf{41.22}&	\textbf{39.29}&	\textbf{39.46}	&\textbf{60.23}&	\textbf{60.08}&	\textbf{60.54}\\
  \bottomrule 
  \end{tabularx}
  \label{dair}
\end{table*}

%% file: latex/fig/distance_correlation.tex
\begin{figure}[t!]
	\centering
	\includegraphics[width=8.5cm]{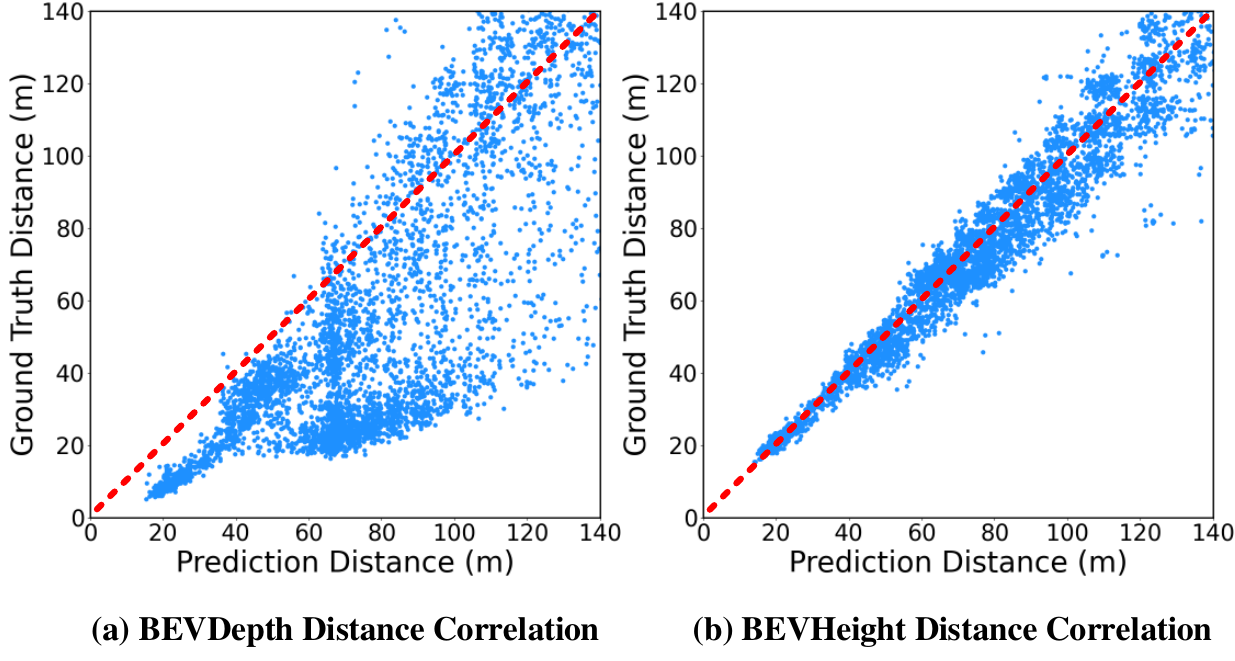}
	\caption{\textbf{Empirical analysis of the distance correlation.} All experiments are conducted on the DAIR-V2X-I val set. (a) and (b) reveal the distance correlation between ground truth and predicted distance on the BEVDepth and our BEVHeight. We take distances from the camera's coordinate system origin to the annotated objects' center for consideration. Each point represents an annotated instance. The scatter diagram of BEVHeight in (b) is closer to the diagonal than that of BEVDepth in (a), indicating that the distance error triggered by height estimation is more minimal than the depth candidate.}

\label{fig:distance_correlation}
\end{figure}

%% file: latex/table/pc_sup.tex
\begin{table}[t]
 \scriptsize\centering\addtolength{\tabcolsep}{-2.5pt}
 \caption{\textbf{Results with point cloud supervision on DAIR-V2X-I dataset.} We can observe that for both BEVDepth and BEVHeight, LiDAR point cloud supervision did not help in terms of evaluation results. This is another evidence that road-side perception is different from the ego-vehicle one.  }
 \begin{tabularx}{1.0\linewidth}{l|ccc|ccc|ccc}
  \toprule
 \multirow{2}{*}{Method}    & \multicolumn{3}{c|}{$\text{Veh.}_{(IoU=0.5)}$} & \multicolumn{3}{c|}{$\text{Ped.}_{(IoU=0.25)}$} & \multicolumn{3}{c}{$\text{Cyc.}_{(IoU=0.25)}$}  \\
   \cmidrule(r){2-10}
 & Easy & Mid & Hard & Easy & Mid & Hard & Easy & Mid & Hard  
 \\
    \midrule
    BEVDepth	& 71.56& 	60.75&	60.85&	21.55&	20.51&	20.75&	40.83	&40.66&	40.26\\
    BEVDepth$\dagger$	&	71.09&	60.37&	60.46&	21.23&	20.84&	20.85&	40.54&	40.34&	40.32\\
    \midrule
    BEVHeight	 &	75.58&	63.49&	63.59&	26.93&	25.47&	\textbf{25.78}&	47.97	& 47.45	& 48.12	\\
    BEVHeight$\dagger$	& \textbf{75.64}& \textbf{63.61}&	\textbf{63.72}&	\textbf{27.01}&	\textbf{25.55}&	25.34&	\textbf{48.03}&	\textbf{47.62}&	\textbf{48.19}\\
    \bottomrule
    \multicolumn{10}{l}{\scriptsize{$\dagger$ denotes training with PointCloud supervision.}}
  \end{tabularx}
  \label{pc_sup}
\end{table}

%% file: latex/table/ablation_accumulate_feature_fusion.tex
\begin{table}[h!t]
\centering
\caption{\extend{\textbf{Effect of image-view fusion and bird's-eye-view fusion in the feature fusion process.}}}
\begin{tabular}{c|c|ccc}
\toprule\noalign{\smallskip}
\renewcommand\arraystretch{1.20}
\textbf{Image-view Fusion} & \textbf{BEV Fusion} & \textbf{NDS$\uparrow$} & \textbf{mAP$\uparrow$} & \textbf{mAVE$\downarrow$}\\
\noalign{\smallskip}
\hline
\noalign{\smallskip}
$\times$   & $\times$  & 0.475 & 0.351   & 0.428\\
$\times$ & $\checkmark$ & 0.489  & 0.367 & 0.388 \\
$\checkmark$ & $\times$ & 0.492 & 0.369 & 0.378 \\
$\checkmark$ & $\checkmark$ & \textbf{0.498} & \textbf{0.373} & \textbf{0.375}  \\
\bottomrule
\end{tabular}
\label{tab:ablation_acculate_feature_fusion}
\arrayrulecolor{black}
\end{table}

%% file: latex/table/ablation_image_view_fusion.tex
\begin{table}[h!t]
\centering
\addtolength{\tabcolsep}{-2.0pt}
\caption{\extend{
\textbf{Ablation studies on the combination of features in the image-view fusion process.} `Hgt. Dist.': height distribution, `Dep. Dist.': depth distribution.}}
\begin{tabular}{c|c|c|ccc}
\toprule\noalign{\smallskip}
\renewcommand\arraystretch{1.20}
\textbf{Image Features} & \textbf{Dep. Dist.} & \textbf{Hgt. Dist.} & \textbf{NDS$\uparrow$} & \textbf{mAP$\uparrow$} & \textbf{mAVE$\downarrow$}\\
\noalign{\smallskip}
\hline
\noalign{\smallskip}
$\checkmark$   & $\times$   & $\times$ & 0.489     & 0.367 & 0.388\\
$\checkmark$ & $\checkmark$ & $\times$ & 0.467     & 0.344 & 0.432  \\
$\checkmark$ & $\times$ & $\checkmark$ & \textbf{0.498}  & \textbf{0.373} & \textbf{0.375}  \\
$\checkmark$ & $\checkmark$ & $\checkmark$ &  0.476 & 0.350 & 0.415  \\
\bottomrule
\end{tabular}
\label{tab:ablation_image_view_fusion}
\arrayrulecolor{black}
\end{table}

%% file: latex/table/ablation_sampled_features.tex
\begin{table}[h!t]
\centering
\caption{\extend{\textbf{Ablation studies on the selection of sampled features at reference points in the bird's-eye-view process.} $F_{H}^{bev}$ denotes height-based BEV features, $F_{D}^{bev}$ denotes depth-based BEV features.}}
\begin{tabular}{c|ccc|ccc}
\toprule
\renewcommand\arraystretch{1.20}
\multirow{2}{*}{\textbf{$Q'_{p}$}}& \multicolumn{3}{c|}{\textbf{nuScenes}} & \multicolumn{3}{c}{\textbf{DAIR-V2X-I}} \\
\cmidrule(r){2-7}
~& \textbf{NDS$\uparrow$} & \textbf{mAP$\uparrow$} & \textbf{mAVE$\downarrow$} & \textbf{Easy} & \textbf{Moderate} & \textbf{Hard} \\
\noalign{\smallskip}
\hline
\noalign{\smallskip}
$F_{H}^{bev}$ & 0.492 & 0.371 & 0.379 & \textbf{79.31} & \textbf{68.62} & \textbf{68.68}\\
\noalign{\smallskip}
$F_{D}^{bev}$ & \textbf{0.498}  & \textbf{0.373} & \textbf{0.375} & 78.95 & 67.26 & 67.36\\
\bottomrule
\end{tabular}
\label{tab:ablation_sampled_features}
\arrayrulecolor{black}
\end{table}

%% file: latex/table/ablation_multi_stage_training.tex
\begin{table}[h!t]
\centering
\addtolength{\tabcolsep}{1.8pt}
\caption{\extend{\textbf{Ablation studies on the effect of the multi-stage training strategy.
}}}
\begin{tabular}{c|c|ccc}
\toprule\noalign{\smallskip}
\renewcommand\arraystretch{2.0}
~& \textbf{Multi-stage Training} & \textbf{NDS$\uparrow$} & \textbf{mAP$\uparrow$} & \textbf{mAVE$\downarrow$}\\
\noalign{\smallskip}
\hline
\noalign{\smallskip}
(a) & w/o  &   0.492   & 0.364 & 0.377\\
(b) & depth branch \textbf{$\rightarrow$} others & \textbf{0.498}  & 0.373 & \textbf{0.375}  \\
(c) & height branch \textbf{$\rightarrow$} others & 0.497  & \textbf{0.374} & 0.379  \\
\bottomrule
\end{tabular}
\label{tab:multi_training_stage}
\arrayrulecolor{black}
\end{table}

%% file: latex/table/ablation_versatility.tex
\begin{table}[h!t]
\centering
\addtolength{\tabcolsep}{-0.5pt}
\caption{\extend{\textbf{Ablation studies on different depth-based meth-
ods.} Here, we conduct the evaluation on the nuScenes val
set. ‘D’ represents depth-based vision transformation, ‘DH’ implies the one fusing depth and height.
}}
\begin{tabular}{l|c|ccc}
\toprule\noalign{\smallskip}
\renewcommand\arraystretch{1.0}
\textbf{Method}& \textbf{View Transformation} & \textbf{NDS$\uparrow$} & \textbf{mAP$\uparrow$} & \textbf{mAVE$\downarrow$}\\
\noalign{\smallskip}
\hline
\noalign{\smallskip}

\multirow{2}{*}{BEVDet~\cite{huang2021bevdet}} & D & 0.453 & 0.323 & 0.429  \\
~ & DH & 0.474    & 0.346 & 0.402  \\

\noalign{\smallskip}
\hline
\noalign{\smallskip}

\multirow{2}{*}{BEVDepth\cite{li2022bevdepth}} & D & 0.475 & 0.351 & 0.428\\
~ & DH & 0.498  & 0.373 & 0.375  \\

\bottomrule
\end{tabular}
\label{tab:ablation_versatility}
\arrayrulecolor{black}
\end{table}